\documentclass{article}

\PassOptionsToPackage{numbers, compress}{natbib}

\usepackage[preprint]{neurips_2023}

\usepackage[utf8]{inputenc} 
\usepackage[T1]{fontenc}    
\usepackage[pagebackref=false,breaklinks,colorlinks]{hyperref}       
\usepackage{booktabs}       
\usepackage{amsfonts}       
\usepackage{amssymb}
\usepackage{amsmath}
\newcommand{\todo}[1]{{\color{red}#1}}
\usepackage{enumitem}
\usepackage{graphicx} 
\usepackage{subcaption}
\usepackage{multirow}
\usepackage{multicol}
\usepackage[table,xcdraw]{xcolor}
\usepackage{wrapfig}
\usepackage[normalem]{ulem}
\useunder{\uline}{\ul}{}

\title{Are We on the Right Way for Evaluating \\ Large Vision-Language Models?}

\author{
{Lin Chen{$^{1,3}$\thanks{Equal contribution. This work is done during internship in Shanghai AI Laboratory.}} ~ Jinsong Li{$^{2,3*}$} ~ Xiaoyi Dong{$^{2,3}$} ~ Pan Zhang{$^{3}$} ~ Yuhang Zang{$^{3}$}} \\ \textbf{Zehui Chen{$^{1}$}~Haodong Duan{$^{3}$}~Jiaqi Wang{$^{3}$\thanks{Correspoding author.}}~~Yu Qiao{$^{3}$}~Dahua Lin{$^{2,3}$}~ Feng Zhao{$^{1\dag}$}} \\
\normalsize
$^{1}$\	University of Science and Technology of China \\ 
$^{2}$\ The Chinese University of Hong Kong\\
$^{3}$\ Shanghai AI Laboratory\\
\url{https://mmstar-benchmark.github.io/} \\
}

\begin{document}

\maketitle

\begin{abstract}
Large vision-language models (LVLMs) have recently achieved rapid progress, sparking numerous studies to evaluate their multi-modal capabilities.
However, we dig into current evaluation works and identify two primary issues: 
1) \textbf{Visual content is unnecessary for many samples.} The answers can be directly inferred from the questions and options, or the world knowledge embedded in LLMs. This phenomenon is prevalent across current benchmarks. For instance, GeminiPro achieves 42.9\% on the MMMU benchmark \textit{without} any visual input, and outperforms the random choice baseline across six benchmarks over 24\% on average.
2) \textbf{Unintentional data leakage exists in LLM and LVLM training.} LLM and LVLM could still answer some visual-necessary questions without visual content, indicating the memorizing of these samples within large-scale training data. For example, Sphinx-X-MoE gets 43.6\% on MMMU \textit{without} accessing images, surpassing its LLM backbone with 17.9\%.
Both problems lead to misjudgments of actual multi-modal gains and potentially misguide the study of LVLM. To this end, we present \textbf{MMStar}, an elite vision-indispensable multi-modal benchmark comprising 1,500 samples meticulously selected by humans. MMStar benchmarks 6 core capabilities and 18 detailed axes, aiming to evaluate LVLMs' multi-modal capacities with carefully balanced and purified samples.
These samples are first roughly selected from current benchmarks with an automated pipeline, human review is then involved to ensure each curated sample exhibits visual dependency, minimal data leakage, and requires advanced multi-modal capabilities.
Moreover, two metrics are developed to measure data leakage and actual performance gain in multi-modal training. 
We evaluate 16 leading LVLMs on MMStar to assess their multi-modal capabilities, and on 7 benchmarks with the proposed metrics to investigate their data leakage and actual multi-modal gain. 


\end{abstract}

\section{Introduction}
\label{sec:intro}
Encouraged by the rapid development of large language models (LLMs) \cite{yang2023baichuan,brown2020language,chiang2023vicuna,chowdhery2022palm,du2021glm,bai2023qwen,touvron2023llama}, integrating visual modality into LLMs to enhance models' interactivity capabilities has witnessed ever-changing advances in recent days \cite{zhu2023minigpt,liu2023visual,liu2023improved,instructblip,zhang2023internlm,bai2023qwenvl,ye2023mplug,luo2023cheap, chen2023sharegpt4v,dong2024internlm}. These large vision-language models (LVLMs) showcase powerful visual perception and understanding capabilities, enabling them to accept image inputs from users and engage in dialogues, thereby offering a more enriched interactive experience. These achievements have further inspired the research community to develop a variety of multi-modal benchmarks \cite{li2023seed, fu2023mme, liu2023mmbench, schwenk2022okvqa, yu2023mm, yue2023mmmu, lu2023mathvista, Kembhavi2016ADI, lu2022learn}, constructed to explore the powerful capabilities emerging from LVLMs and provide a comprehensive and objective platform for quantitatively comparing the continually evolving models. Despite the race among existing evaluation works to construct as many axes as possible to assess the capabilities of LVLMs, we have identified two primary issues upon delving into existing evaluation samples and processes.

\begin{figure*}[t]
    \centering
    \includegraphics[width=\linewidth]{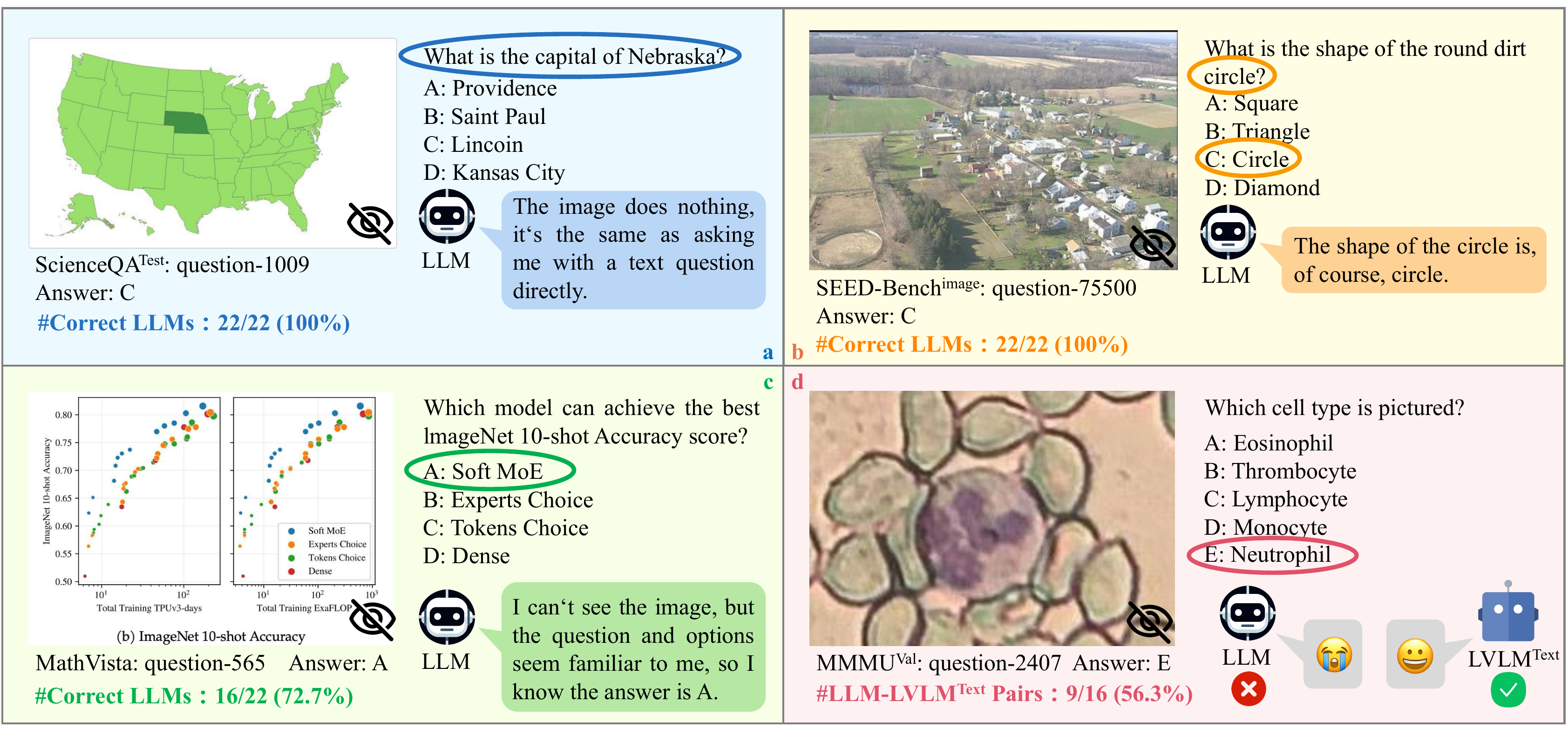}
    \captionsetup{font={footnotesize}}
    \caption{We highlight cases in existing multi-modal benchmarks where evaluation samples either \textbf{lack visual dependency} or \textbf{have unintentionally leaked into the training data of LLMs and LVLMs}. \textbf{(a)} Some samples can be answered by LLMs using only text-based world knowledge; \textbf{(b)} For some instances, the question itself contains the answer, making images superfluous; \textbf{(c)} Some samples are leaked into LLMs' training corpora can be "recalled" with the textual questions and answers directly; \textbf{(d)} Some samples indiscernible to LLMs but solved by LVLMs without accessing images suggest leakage into LVLMs' multi-modal training data.}
    \label{fig:llm_case}
\end{figure*}

First, \textbf{visual content is unnecessary for many samples}. A qualified multi-modal evaluation sample should compel LVLMs to understand and reason with the visual content for correct answers. Otherwise, the evaluation sample would degrade into assessing the textual capabilities of LLM bases. Unfortunately, we have identified numerous samples across multiple popular benchmarks \cite{liu2023mmbench,li2023seed,yue2023mmmu,lu2022learn,Kembhavi2016ADI} where answers can be correctly deduced without relying on visual content. As shown in Figure \ref{fig:llm_case} (a) and (b), some samples have answers directly included within the questions (e.g., What is the shape of the round dirt circle?), while others can be effortlessly answered by leveraging the rich world knowledge embedded within the LLM bases (e.g., What is the capital of Nebraska?). 
With a comprehensive quantitative analysis of 25 LLMs on 6 benchmarks, we observe this phenomenon is prevalent and serious. For example, more than 50\% questions of ScienceQA and 20\% questions of MMMU can be solved by most LLMs directly. For the powerful close source LLM GeminiPro, it achieves 42.9\% on the MMMU benchmark without any visual input, and outperforms the random choice baseline across six benchmarks by over 24\% on average. 

Taking aside the inappropriate samples in evaluation, we also observed strange results that LLM and LVLM could still answer some visual-necessary questions without visual content (Figure \ref{fig:llm_case} (c) and (d)). A plausible explanation for this could be the inadvertent memorization of these samples during the large-scale training process, suggesting the presence of \textbf{unintentional data leakage in the training of LLM and LVLM}.
Through a detailed study of 22 LVLMs on the 6 benchmarks, we find the unexpected leaking problem during the LVLM training is particularly serious. For example, we find Yi-VL-34B gets $15.0\%$ higher performance than its LLM backbone on ScienceQA, Sphinx-X-MoE gets $43.6\%$ on MMMU \textit{without} accessing images, surpassing its LLM backbone with $17.9\%$, even surpassing many leading LVLMs with accessing images.

The existence of inappropriate questions and data leaking would lead to misjudgments of actual multi-modal performance gains and potentially misguide the study of LVLM. 
In pursuit of a more accurate and comprehensive evaluation, we introduce the MMStar Benchmark. MMStar is a premier, vision-critical multi-modal benchmark that includes 1,500 challenging samples, each rigorously validated by humans. It is structured to test 6 fundamental capabilities and 18 specific dimensions, aiming to evaluate the multi-modal capacities of LVLMs with a carefully balanced and purified selection of samples.

The MMStar is a new benchmark that ``Stands on the shoulders of giants''. Samples are first roughly selected from current benchmarks with an automated pipeline.
In detail, we use eight powerful LLMs as candidates inspectors for visual dependency and LLM leakage, including two closed-source APIs (GPT4-Turbo \cite{chatgpt}, and GeminiPro \cite{team2023gemini}) and six leading open-source models (e.g., LLaMA-70B \cite{touvron2023llama}, Qwen-1.5-72B \cite{bai2023qwen}, and Mixtral-8x7B \cite{jiang2024mixtral}). 
Samples that could be answered by more than 2 of the 8 LLMs are excluded as they may exist leaking or visual-unnecessary problems. 
Then we use 16 leading LVLMs (e.g., GPT4V \cite{gpt4v}, GeminiPro-Vision \cite{team2023gemini}, LLaVA series \cite{liu2023improved, liu2023visual}) to gauge the difficulty of the samples and split them to four levels. Ultimately, based on the difficulty of the rough-filtered samples, \textbf{strict manual review and selection} are applied to curate 1,500 high-quality multimodal evaluation samples. As shown in Figure \ref{fig:mmstar_axes}, these samples span 6 core multimodal capability dimensions and 18 detailed axes, aiming to probe LVLMs' advanced multimodal capabilities with a purified and high-quality set of samples. Moreover, we design the multi-modal gain (MG) and multi-modal leakage (ML) metrics to probe LVLMs' actual performance gain and data leakage degrees derived from multi-modal training in a benchmark-specific manner.

We evaluate the accuracy, MG, and ML of 16 leading LVLMs on our MMStar benchmark, the high-resolution version of GPT-4V ranks first with $57.1\%$ accuracy, showcasing its superb multi-modal capability. GPT-4V also gets the best MG and a small ML, indicating its effective multi-modal training strategy and has less data leaking. 

In a nutshell, our contributions are threefold: 
\begin{itemize}[leftmargin=*,noitemsep,nolistsep]
    \item We delve into existing evaluation benchmarks and processes and identify two key issues: (1) Visual content is unnecessary for many samples.  (2) Unintentional data leakage exists in LLM and LVLM training. Both lead to misjudgment of LVLM capability and may misguide the following study.

    \item We curate MMStar, an elite vision-indispensable multi-modal benchmark comprising 1,500 challenge samples meticulously selected by humans. MMStar covers samples from diverse tasks and difficulties, aiming to evaluate the actual multi-modal capacities of LVLMs.

    \item Based on MMStar, we evaluate LVLMs with Accuracy and two newly proposed metrics: multi-modal gain and multi-modal leakage.  The high-resolution version of GPT-4V outperforms the 16 leading LLMs and ranks first.
\end{itemize}

\section{Related Work}
\textbf{Large Vision-Language Models.} As large language models (LLMs) \cite{chiang2023vicuna,touvron2023llama,touvron2023llama,yang2023baichuan,team2023internlm, chatgpt, ouyang2022training, chowdhery2022palm} rapidly advance, a growing fraction of the research community is focusing on integrating visual content into LLMs to build a powerful intelligent assistant with more interactive ways. Central to these large vision-language models (LVLMs) are the seminal works in modality alignment within the vision-language learning area \cite{radford2021learning,jia2021scaling}. The foundation work CLIP \cite{radford2021learning} exemplifies the alignment of vision and language modalities through contrastive learning on extensive image-text pairs. Built upon the CLIP image encoder which is somewhat aligned with the language modality, current LVLMs typically utilize vast image-text pairs to connect the vision encoder and LLM, enabling LLM to receive and understand visual content \cite{zhu2023minigpt,liu2023visual,liu2023improved,instructblip,zhang2023internlm,bai2023qwenvl,ye2023mplug,luo2023cheap, chen2023sharegpt4v}. For example, MiniGPT4 \cite{zhu2023minigpt} and LLaVA \cite{liu2023visual} directly connect the vision encoder and LLM with QFormer \cite{li2023blip} and MLP \cite{taud2018multilayer}, showing proficiency in multi-modal dialogues. Subsequent works have further enhanced LVLMs by improving the multi-modal instruction data \cite{liu2023improved, ye2023mplug, chen2023sharegpt4v, wang2023see} and designing novel modules \cite{bai2023qwenvl, li2023monkey, wang2023cogvlm, lu2024deepseek, gao2024sphinx,dong2024internlm} for more sufficient modality alignment.

\textbf{Evaluations of LVLMs.} To probe the true capabilities of the emerging LVLMs, the research community has developed many multi-modal benchmarks encompassing a wide range of evaluation axes \cite{liu2023mmbench,fu2023mme,schwenk2022okvqa,yue2023mmmu,sharma2018conceptual,li2023seed, liu2023visual, yu2023mm, wu2023q}. Early single-task benchmarks, such as VQA \cite{goyal2017making}, MS-COCO \cite{sharma2018conceptual}, and OK-VQA \cite{schwenk2022okvqa}, fail to holistically assess LVLMs' general multi-modal perception and reasoning capabilities. To address this issue, comprehensive multi-modal benchmarks have been constructed \cite{liu2023visual, li2023seed, yue2023mmmu, fu2023mme, liu2023mmbench, cheng2023can,wu2023q}. For example, SEED \cite{li2023seed} and MMBench \cite{liu2023mmbench} cover 12 and 20 evaluation dimensions respectively, while MMMU \cite{yue2023mmmu} spans 30 college-level subjects, providing some competitive arenas for a comprehensive comparison of cutting-edge LVLMs. However, existing evaluations of LVLMs overlook some critical issues. On the one hand, they do not guarantee that all evaluation samples can not be correctly answered without the visual content. On the other hand, current evaluations consistently adhere to the process of inferring on given benchmarks and calculating scores for LVLMs, overlooking the possibility of data leakage during multi-modal training. This oversight can lead to unfair comparisons and misjudgments of the real gains in multi-modal capabilities brought by multi-modal training.

\renewcommand{\arraystretch}{1.1}
\begin{table*}[!t]
    \centering
    \footnotesize
    \captionsetup{font={footnotesize}}
    \caption {\textbf{Evaluation of various LLMs on six popular multi-modal benchmarks.} We employ a 0-shot inference strategy for evaluating all LLMs. We report the results of 2 closed-source LLMs and 20 open-source LLMs with varying sizes and architectures. The evaluated benchmarks include MMMU (MMMU-Val \cite{yue2023mmmu}), MMB (MMBench-EN-Dev \cite{liu2023mmbench}), ScienceQA (ScienceQA-Test \cite{lu2022learn}), AI2D (AI2D-Test \cite{Kembhavi2016ADI}), SEED (SEED-Image \cite{li2023seed}), and MathVista (MathVista-Mini \cite{lu2023mathvista}). The {\ul\textbf{best}} results are highlighted in {\ul \textbf{bold and underlined.}}}
    \label{tab:llm}
    \scalebox{0.96}{
    \begin{tabular}{lcccccccc}
    \toprule
    \multicolumn{1}{l|}{Model}           & \multicolumn{1}{l|}{Strategy} & MMMU      & MMB    & ScienceQA      & AI2D     & SEED      & \multicolumn{1}{l|}{MathVista}     & Avg.          \\ \midrule
    \multicolumn{9}{c}{\textit{Baselines}}                                                                                                                                                     \\ \midrule
    \multicolumn{1}{l|}{Random Choice}   & \multicolumn{1}{l|}{-}             & 22.1          & 0.0          & 24.2          & 23.8          & 24.3          & \multicolumn{1}{l|}{17.9}          & 18.7          \\ \midrule
    \multicolumn{9}{c}{\textit{Closed-source LLMs}}                                                                                                                                           \\ \midrule
    \multicolumn{1}{l|}{GPT4-Turbo\cite{chatgpt}}      & \multicolumn{1}{l|}{0-shot}     & 41.2          & 12.2          & 64.3          & {\ul\textbf{59.7}} & 10.1          & \multicolumn{1}{l|}{{\ul\textbf{24.2}}} & 35.3          \\
    \multicolumn{1}{l|}{GeminiPro\cite{team2023gemini}}       & \multicolumn{1}{l|}{0-shot}     & {\ul \textbf{42.9}} & {\ul \textbf{18.4}} & {\ul \textbf{68.9}} & 59.2          & {\ul\textbf{35.5}} & \multicolumn{1}{l|}{23.3}          & {\ul\textbf{41.4}} \\ \midrule
    \multicolumn{9}{c}{\textit{Open-source LLMs}}                                                                                                                                             \\ \midrule
    \multicolumn{1}{l|}{Qwen1.5-1.8B\cite{bai2023qwen}}    & \multicolumn{1}{l|}{0-shot}     & 29.0          & 10.0          & 54.3          & 37.9          & 28.9          & \multicolumn{1}{l|}{20.4}          & 30.1          \\
    \multicolumn{1}{l|}{Phi2-2.7B\cite{phi2}}    & \multicolumn{1}{l|}{0-shot}     & 20.0          & 7.2          & 47.1          & 38.7          & 26.4          & \multicolumn{1}{l|}{22.0}          & 26.9          \\
    \multicolumn{1}{l|}{Yi-6B\cite{young2024yi}}           & \multicolumn{1}{l|}{0-shot}     & 25.7          & 9.5           & 58.1          & 39.1          & 27.4          & \multicolumn{1}{l|}{21.2}          & 30.2          \\
    \multicolumn{1}{l|}{LLaMA2-7B\cite{touvron2023llama}}       & \multicolumn{1}{l|}{0-shot}     & 23.6          & 11.5          & 56.8          & 43.5          & 31.7          & \multicolumn{1}{l|}{24.1}          & 31.9          \\
    \multicolumn{1}{l|}{Qwen-7B\cite{bai2023qwen}}         & \multicolumn{1}{l|}{0-shot}     & 19.8          & 8.4           & 52.7          & 42.6          & 7.6           & \multicolumn{1}{l|}{20.5}          & 25.3          \\
    \multicolumn{1}{l|}{Deepseek-7B\cite{bi2024deepseek}}     & \multicolumn{1}{l|}{0-shot}     & 21.6          & 8.4           & 56.3          & 38.1          & 13.4          & \multicolumn{1}{l|}{20.6}          & 26.4          \\
    \multicolumn{1}{l|}{InternLM2-7B\cite{team2023internlm}}    & \multicolumn{1}{l|}{0-shot}     & 32.8          & 8.9           & 64.0          & 48.3          & 31.9          & \multicolumn{1}{l|}{18.9}          & 34.1          \\
    \multicolumn{1}{l|}{Qwen1.5-7B\cite{bai2023qwen}}      & \multicolumn{1}{l|}{0-shot}     & 25.0          & 11.4          & 62.3          & 49.4          & 19.4          & \multicolumn{1}{l|}{19.9}          & 31.2          \\
    \multicolumn{1}{l|}{Vicuna-v1.5-7B\cite{chiang2023vicuna}}  & \multicolumn{1}{l|}{0-shot}     & 29.9          & 10.3          & 58.9          & 42.5          & 32.6          & \multicolumn{1}{l|}{22.0}          & 32.7          \\
    \multicolumn{1}{l|}{Baichuan2-7B\cite{yang2023baichuan}}    & \multicolumn{1}{l|}{0-shot}     & 25.7          & 10.5          & 52.7          & 44.0          & 29.2          & \multicolumn{1}{l|}{20.8}          & 30.5          \\
    \multicolumn{1}{l|}{Mistral-7B\cite{jiang2023mistral}}      & \multicolumn{1}{l|}{0-shot}     & 30.0          & 13.2          & 63.4          & 48.5          & 34.3          & \multicolumn{1}{l|}{22.6}          & 35.3          \\
    \multicolumn{1}{l|}{LLaMA2-13B\cite{touvron2023llama}}      & \multicolumn{1}{l|}{0-shot}     & 24.4          & 10.1          & 59.1          & 45.0          & 33.6          & \multicolumn{1}{l|}{23.8}          & 32.7          \\
    \multicolumn{1}{l|}{Vicuna-v1.5-13B\cite{chiang2023vicuna}} & \multicolumn{1}{l|}{0-shot}     & 28.3          & 11.6          & 59.5          & 45.0          & 26.3          & \multicolumn{1}{l|}{19.6}          & 31.7          \\
    \multicolumn{1}{l|}{Baichuan2-13B\cite{yang2023baichuan}}   & \multicolumn{1}{l|}{0-shot}     & 22.1          & 4.7           & 51.1          & 32.8          & 25.4          & \multicolumn{1}{l|}{20.3}          & 26.1          \\
    \multicolumn{1}{l|}{InternLM2-20B\cite{team2023internlm}}   & \multicolumn{1}{l|}{0-shot}     & 32.2          & {\ul\textbf{15.9}} & 63.8          & 55.7          & 26.0          & \multicolumn{1}{l|}{21.3}          & 35.8          \\
    \multicolumn{1}{l|}{Yi-34B\cite{young2024yi}}          & \multicolumn{1}{l|}{0-shot}     & {\ul\textbf{37.1}} & 10.5          & 53.6          & 57.3          & {\ul\textbf{37.3}} & \multicolumn{1}{l|}{21.7}          & {\ul\textbf{36.3}} \\
    \multicolumn{1}{l|}{Mixtral-8x7B\cite{jiang2024mixtral}}    & \multicolumn{1}{l|}{0-shot}     & 25.7          & 8.6           & 57.2          & 48.7          & 13.5          & \multicolumn{1}{l|}{23.4}          & 29.5          \\
    \multicolumn{1}{l|}{Deepseek-67B\cite{bi2024deepseek}}    & \multicolumn{1}{l|}{0-shot}     & 30.9          & 14.8          & {\ul\textbf{64.3}} & {\ul\textbf{57.5}} & 17.1          & \multicolumn{1}{l|}{23.2}          & 34.6          \\
    \multicolumn{1}{l|}{LLaMA2-70B\cite{touvron2023llama}}      & \multicolumn{1}{l|}{0-shot}     & 28.9          & 12.3          & 62.2          & 48.6          & 34.3          & \multicolumn{1}{l|}{{\ul\textbf{25.2}}} & 35.3          \\
    \multicolumn{1}{l|}{Qwen1.5-72B\cite{bai2023qwen}}     & \multicolumn{1}{l|}{0-shot}     & 21.4          & 10.1          & 57.5          & 44.2          & 8.8           & \multicolumn{1}{l|}{19.5}          & 26.9          \\ \bottomrule
    \end{tabular}
}
\end{table*}

\section{Two Overlooked Issues for Evaluating LVLMs}
\label{sec:issues}
In this section, we delve into two commonly overlooked issues in current LVLM evaluation works. Moreover, we present detailed experimental results to further substantiate our observations.

\renewcommand{\arraystretch}{1.1}
\begin{table*}[!t]
    \centering
    \footnotesize
    \captionsetup{font={footnotesize}}
    \caption {\textbf{Evaluation of various LLMs on six popular multi-modal benchmarks under 2-shot.} We employ a 2-shot inference strategy for evaluating all LLMs to reduce instances of refusal to answer and align the answer formats. We report the results of 2 closed-source LLMs and 20 open-source LLMs with varying sizes and architectures. The evaluated benchmarks include MMMU (MMMU-Val \cite{yue2023mmmu}), MMB (MMBench-EN-Dev \cite{liu2023mmbench}), ScienceQA (ScienceQA-Test \cite{lu2022learn}), AI2D (AI2D-Test \cite{Kembhavi2016ADI}), SEED (SEED-Image \cite{li2023seed}), and MathVista (MathVista-Mini \cite{lu2023mathvista}). The {\ul \textbf{best}} results are highlighted in {\ul \textbf{bold and underlined.}}}
    \label{tab:llm_2_shot}
    \scalebox{0.96}{
    \begin{tabular}{lcccccccc}
    \toprule
    \multicolumn{1}{l|}{Model}           & \multicolumn{1}{l|}{Strategy} & MMMU      & MMB    & ScienceQA      & AI2D     & SEED      & \multicolumn{1}{l|}{MathVista}     & Avg.          \\ \midrule
    \multicolumn{9}{c}{\textit{Baseline}}                                                                                                                                                     \\ \midrule
    \multicolumn{1}{l|}{Random Choice}   & \multicolumn{1}{l|}{-}             & 22.1          & 0.0          & 24.2          & 23.8          & 24.3          & \multicolumn{1}{l|}{17.9}          & 18.7          \\ \midrule
    \multicolumn{9}{c}{\textit{Closed-source LLMs}}                                                                                                                                           \\ \midrule
    \multicolumn{1}{l|}{GPT4-Turbo\cite{chatgpt}}      & \multicolumn{1}{l|}{2-shot}     & 42.0          & 15.5          & 67.5          & {\ul\textbf{61.3}} & 26.8          & \multicolumn{1}{l|}{{\ul\textbf{25.6}}} & 39.8          \\
    
    \multicolumn{1}{l|}{GeminiPro\cite{team2023gemini}}       & \multicolumn{1}{l|}{2-shot}     & {\ul\textbf{42.7}} & {\ul\textbf{18.7}} & {\ul\textbf{69.3}} & 60.1          & {\ul\textbf{38.1}} & \multicolumn{1}{l|}{25.5}          & {\ul\textbf{42.4}} \\ \midrule
    
    \multicolumn{9}{c}{\textit{Open-source LLMs}}                                                                                                                                   \\ \midrule
    \multicolumn{1}{l|}{Qwen1.5-1.8B\cite{bai2023qwen}}    & \multicolumn{1}{l|}{2-shot}        & 33.0          & 8.6           & 55.6          & 41.3          & 32.1          & \multicolumn{1}{l|}{22.7}          & 32.2          \\
    \multicolumn{1}{l|}{Phi2-2.7B\cite{phi2}}    & \multicolumn{1}{l|}{2-shot}     & 19.9          & 4.3          & 50.8          & 41.7          & 6.9          & \multicolumn{1}{l|}{18.4}          & 23.7          \\
    \multicolumn{1}{l|}{Yi-6B\cite{young2024yi}}           & \multicolumn{1}{l|}{2-shot}        & 32.9          & 16.0          & 64.6          & 51.5          & 36.7          & \multicolumn{1}{l|}{24.5}          & 37.7          \\
    \multicolumn{1}{l|}{LLaMA2-7B\cite{touvron2023llama}}       & \multicolumn{1}{l|}{2-shot}        & 25.9          & 7.7           & 57.9          & 42.8          & 32.8          & \multicolumn{1}{l|}{22.8}          & 31.7          \\
    \multicolumn{1}{l|}{Qwen-7B\cite{bai2023qwen}}         & \multicolumn{1}{l|}{2-shot}        & 30.6          & 15.0          & 63.0          & 50.0          & 32.6          & \multicolumn{1}{l|}{21.0}          & 35.4          \\
    \multicolumn{1}{l|}{Deepseek-7B\cite{bi2024deepseek}}     & \multicolumn{1}{l|}{2-shot}        & 28.7          & 11.6          & 61.9          & 46.0          & 34.1          & \multicolumn{1}{l|}{21.7}          & 34.0          \\
    \multicolumn{1}{l|}{InternLM2-7B\cite{team2023internlm}}    & \multicolumn{1}{l|}{2-shot}        & 33.6          & 11.4          & 63.6          & 52.1          & 34.4          & \multicolumn{1}{l|}{20.4}          & 35.9          \\
    \multicolumn{1}{l|}{Qwen1.5-7B\cite{bai2023qwen}}      & \multicolumn{1}{l|}{2-shot}        & 33.3          & 13.1          & 65.1          & 52.1          & 32.1          & \multicolumn{1}{l|}{22.8}          & 36.4          \\
    \multicolumn{1}{l|}{Vicuna-v1.5-7B\cite{chiang2023vicuna}}  & \multicolumn{1}{l|}{2-shot}        & 31.3          & 9.5           & 58.9          & 45.5          & 32.0          & \multicolumn{1}{l|}{20.7}          & 33.0          \\
    \multicolumn{1}{l|}{Baichuan2-7B\cite{yang2023baichuan}}    & \multicolumn{1}{l|}{2-shot}        & 28.2          & 13.7          & 58.1          & 44.1          & 32.3          & \multicolumn{1}{l|}{21.7}          & 33.0          \\
    \multicolumn{1}{l|}{Mistral-7B\cite{jiang2023mistral}}      & \multicolumn{1}{l|}{2-shot}        & 29.8          & 17.2          & 66.1          & 50.0          & 34.4          & \multicolumn{1}{l|}{13.4}          & 35.2          \\
    \multicolumn{1}{l|}{LLaMA2-13B\cite{touvron2023llama}}      & \multicolumn{1}{l|}{2-shot}        & 32.9          & 10.1          & 58.9          & 43.8          & 32.1          & \multicolumn{1}{l|}{24.8}          & 33.8          \\
    \multicolumn{1}{l|}{Vicuna-v1.5-13B\cite{chiang2023vicuna}} & \multicolumn{1}{l|}{2-shot}        & 31.3          & 12.8          & 63.0          & 46.8          & 33.6          & \multicolumn{1}{l|}{20.8}          & 34.7          \\
    \multicolumn{1}{l|}{Baichuan2-13B\cite{yang2023baichuan}}   & \multicolumn{1}{l|}{2-shot}        & 32.2          & 13.1          & 61.0          & 47.1          & 35.2          & \multicolumn{1}{l|}{23.4}          & 35.3          \\
    \multicolumn{1}{l|}{InternLM2-20B\cite{team2023internlm}}   & \multicolumn{1}{l|}{2-shot}        & 35.6          & 17.4          & 66.4          & 55.9          & 30.4          & \multicolumn{1}{l|}{20.8}          & 37.8          \\
    \multicolumn{1}{l|}{Yi-34B\cite{young2024yi}}          & \multicolumn{1}{l|}{2-shot}        & 35.8          & 15.8          & 67.9          & 59.6          & 37.2          & \multicolumn{1}{l|}{{\ul\textbf{26.9}}} & 40.5          \\
    \multicolumn{1}{l|}{Mixtral-8x7B\cite{jiang2024mixtral}}    & \multicolumn{1}{l|}{2-shot}        & 35.1          & 17.3          & 66.3          & 55.1          & 35.8          & \multicolumn{1}{l|}{22.7}          & 38.7          \\
    \multicolumn{1}{l|}{Deepseek-67B\cite{bi2024deepseek}}    & \multicolumn{1}{l|}{2-shot}        & 38.3          & 17.2          & 68.3          & 59.7          & 37.3          & \multicolumn{1}{l|}{23.4}          & 40.7          \\
    \multicolumn{1}{l|}{LLaMA2-70B\cite{touvron2023llama}}      & \multicolumn{1}{l|}{2-shot}        & 30.4          & 17.2          & 63.4          & 49.3          & 34.9          & \multicolumn{1}{l|}{24.2}          & 36.6          \\
    \multicolumn{1}{l|}{Qwen1.5-72B\cite{bai2023qwen}}     & \multicolumn{1}{l|}{2-shot}        & {\ul\textbf{42.4}} & {\ul\textbf{21.1}} & {\ul\textbf{70.1}} & {\ul\textbf{60.9}} & {\ul\textbf{40.7}} & \multicolumn{1}{l|}{26.3}          & {\ul\textbf{43.6}} \\ \bottomrule
    \end{tabular}
}
\end{table*}

\textbf{First issue: visual content is unnecessary for many evaluation samples.} The key distinction between evaluating LLMs and LVLMs lies in the necessity for LVLM evaluations to strictly ensure that the correct answers can only be derived based on a thorough understanding of visual content. Without this, evaluating LVLMs' multi-modal capabilities degrades to merely assessing their LLM backbones' uni-modal abilities. However, upon examining samples from some popular LVLM benchmarks, we find many samples lack vital visual dependency and can yield correct answers even without the image inputs!
Through analysis of these failure samples, we categorize them into two groups: (1) Answers can be directly obtained from the world knowledge embedded in LLMs, owing to the LLMs' extensive pertaining on the large corpus of data. For example, as illustrated in Figure \ref{fig:llm_case}(a), the question "What is the capital of Nebraska?" already provides the key information "Nebraska", eliminating the need for extracting relevant location information from visual content. A more appropriate question is "What is the capital of the highlighted area in the image?" to emphasize the importance of visual understanding.
(2) Answers are directly included in the textual questions. As shown in Figure \ref{fig:llm_case}(b), LLMs can derive the correct answer "circle" through simple reasoning based on the question "What is the shape of the round dirt circle?".

\begin{wrapfigure}{r}{0.49\textwidth}
    \vspace{-12pt}
    \centering
    \includegraphics[width=0.48\textwidth]{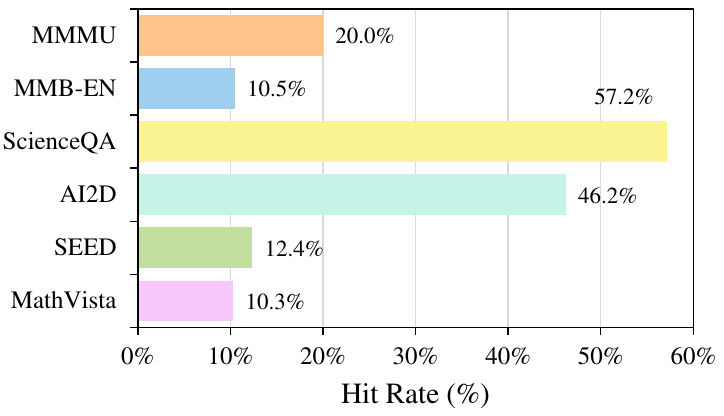}
    \captionsetup{font={footnotesize}}
    \vspace{-3pt}
    \caption{
    LLM hit rate across various benchmarks.
    }
    \label{fig:hit_rate}
    \vspace{-10pt}
\end{wrapfigure}
To quantitatively substantiate our findings, we further experiment to gauge the proportion of these two types of samples in existing benchmarks. Specifically, we evaluate several benchmarks with two closed-source LLMs (GPT4-Turbo \cite{chatgpt}, and GeminiPro \cite{team2023gemini}) and six open-source heavy LLMs (InternLM2-20B \cite{team2023internlm}, Yi-34B \cite{young2024yi}, Mixtral-8x7B \cite{jiang2024mixtral}, Deepseek-67B 
 \cite{bi2024deepseek}, LLaMA2-70B \cite{touvron2023llama}, and Qwen1.5-72B \cite{bai2023qwen}), recording the hit count for each question.
Here, the `hit' refers to the ability of an LLM to correctly answer the question without relying on visual input.
We then calculate the percentage of samples with a hit count of six or more (representing 80\%) against the total number of samples to determine the abnormal hit rate for each benchmark. As depicted in Figure 
\ref{fig:hit_rate}
, every benchmark shows a certain degree of samples that visual contents are unnecessary, with ScienceQA~\cite{lu2022learn} and AI2D~\cite{Kembhavi2016ADI} exhibiting amazing abnormal hit rates of 57.2\% and 46.2\%, respectively. Based on our observations, most multi-modal benchmarks have yet to fully assess the multi-modal capabilities of LVLMs.

\renewcommand{\arraystretch}{1.1}
\begin{table*}[!t]
    \centering
    \footnotesize
    \captionsetup{font={footnotesize}}
    \caption {\textbf{Evaluation of various LVLMs on six popular multi-modal benchmarks.} For the "strategy" column, "LLM" refers to evaluating using the corresponding LLM base of the LVLM, while "LVLM-text" denotes evaluating LVLMs without accessing images. We employ the 0-shot inference strategy for LLMs to align the evaluation protocols of LVLMs. We only report the results of 2 closed-source LVLMs and 8 open-source LVLMs due to space limits. For the entire LVLMs' results, please refer to the appendix. The {\ul\textbf{highest}} results of the LVLM-text setting across the models are highlighted in {\ul \textbf{bold and underlined.}}}
    \label{tab:lvlm}
    \scalebox{0.84}{

\begin{tabular}{lllccccccc}
\toprule
\multicolumn{1}{l|}{Model}                                                                                                               & \multicolumn{1}{l|}{Param.}                                        & \multicolumn{1}{l|}{Strategy}                          & MMMU                & MMB                 & ScienceQA           & AI2D                & SEED                & \multicolumn{1}{c|}{MathVista}                                   & Avg.                \\ \midrule
\multicolumn{10}{c}{\textit{Baseline}}                                                                                                                                                                                                                                                                                                                                                                                                                                        \\ \midrule
\multicolumn{1}{l|}{Random Choice}                                                                                                       & \multicolumn{1}{l|}{-}                                             & \multicolumn{1}{l|}{-}                                 & 22.1                & 0.0                & 24.2                & 23.8                & 24.3                & \multicolumn{1}{c|}{17.9}                                        & 18.7                \\ \midrule
\multicolumn{10}{c}{\textit{Closed-source LVLMs and corresponding LLM bases}}                                                                                                                                                                                                                                                                                                                                                                                                 \\ \midrule
\rowcolor[HTML]{E4E4E4} 
\multicolumn{1}{l|}{\cellcolor[HTML]{E4E4E4}}                                                                                            & \multicolumn{1}{l|}{\cellcolor[HTML]{E4E4E4}-}                     & \multicolumn{1}{l|}{\cellcolor[HTML]{E4E4E4}LLM}       & 41.2                & 12.2                & 64.3                & 59.7                & 10.1                & \multicolumn{1}{c|}{\cellcolor[HTML]{E4E4E4}24.2}                & 35.3                \\
\rowcolor[HTML]{E4E4E4} 
\multicolumn{1}{l|}{\cellcolor[HTML]{E4E4E4}}                                                                                            & \multicolumn{1}{l|}{\cellcolor[HTML]{E4E4E4}-}                     & \multicolumn{1}{l|}{\cellcolor[HTML]{E4E4E4}LVLM-text}    & {\ul \textbf{45.1}} & {\ul \textbf{17.6}} & {\ul \textbf{68.2}} & {\ul \textbf{62.5}} & {\ul \textbf{28.4}} & \multicolumn{1}{c|}{\cellcolor[HTML]{E4E4E4}{\ul \textbf{25.4}}} & {\ul \textbf{41.2}} \\
\rowcolor[HTML]{E4E4E4} 
\multicolumn{1}{l|}{\multirow{-3}{*}{\cellcolor[HTML]{E4E4E4}\begin{tabular}[c]{@{}l@{}}GPT4V\cite{gpt4v}\\ (GPT4-Turbo\cite{chatgpt})\end{tabular}}}              & \multicolumn{1}{l|}{\cellcolor[HTML]{E4E4E4}-}                     & \multicolumn{1}{l|}{\cellcolor[HTML]{E4E4E4}LVLM}      & 53.6                & 69.6                & 81.4                & 75.3                & 71.6                & \multicolumn{1}{c|}{\cellcolor[HTML]{E4E4E4}44.7}                & 66.0                \\
\multicolumn{1}{l|}{}                                                                                                                    & \multicolumn{1}{l|}{-}                                             & \multicolumn{1}{l|}{LLM}                               & 42.9                & 18.4                & 68.9                & 59.2                & 35.5                & \multicolumn{1}{c|}{23.3}                                        & 41.4                \\
\multicolumn{1}{l|}{}                                                                                                                    & \multicolumn{1}{l|}{-}                                             & \multicolumn{1}{l|}{LVLM-text}                            & 39.4                & 16.7                & 66.3                & 54.5                & 27.9                & \multicolumn{1}{c|}{24.5}                                        & 38.2                \\
\multicolumn{1}{l|}{\multirow{-3}{*}{\begin{tabular}[c]{@{}l@{}}GeminiPro-Vision\cite{team2023gemini}\\ (GeminiPro\cite{team2023gemini})\end{tabular}}}                            & \multicolumn{1}{l|}{-}                                             & \multicolumn{1}{l|}{LVLM}                              & 44.4                & 68.1                & 80.6                & 68.0                & 64.3                & \multicolumn{1}{c|}{36.0}                                        & 60.2                \\ \midrule
\multicolumn{10}{c}{\textit{Open-source LVLMs and corresponding LLM bases}}                                                                                                                                                                                                                                                                                                                                                                                                   \\ \midrule
\rowcolor[HTML]{E4E4E4} 
\multicolumn{1}{l|}{\cellcolor[HTML]{E4E4E4}}                                                                                            & \multicolumn{1}{l|}{\cellcolor[HTML]{E4E4E4}}                      & \multicolumn{1}{l|}{\cellcolor[HTML]{E4E4E4}LLM}       & 20.0                & 7.2                 & 47.1                & 38.7                & 26.4                & \multicolumn{1}{c|}{\cellcolor[HTML]{E4E4E4}22.0}                & 26.9                \\
\rowcolor[HTML]{E4E4E4} 
\multicolumn{1}{l|}{\cellcolor[HTML]{E4E4E4}}                                                                                            & \multicolumn{1}{l|}{\cellcolor[HTML]{E4E4E4}}                      & \multicolumn{1}{l|}{\cellcolor[HTML]{E4E4E4}LVLM-text} & 30.0                & 21.0                & 62.3                & 51.9                & 37.2                & \multicolumn{1}{c|}{\cellcolor[HTML]{E4E4E4}23.5}                & 37.7                \\
\rowcolor[HTML]{E4E4E4} 
\multicolumn{1}{l|}{\multirow{-3}{*}{\cellcolor[HTML]{E4E4E4}\begin{tabular}[c]{@{}l@{}}TinyLLaVA\cite{zhou2024tinyllava}\\ (Phi2-2.7B\cite{phi2})\end{tabular}}}           & \multicolumn{1}{l|}{\multirow{-3}{*}{\cellcolor[HTML]{E4E4E4}3B}}  & \multicolumn{1}{l|}{\cellcolor[HTML]{E4E4E4}LVLM}      & 36.0                & 66.9                & 69.1                & 62.4                & 70.1                & \multicolumn{1}{c|}{\cellcolor[HTML]{E4E4E4}28.9}                & 55.6                \\
\multicolumn{1}{l|}{}                                                                                                                    & \multicolumn{1}{l|}{}                                              & \multicolumn{1}{l|}{LLM}                               & 29.9                & 10.3                & 58.9                & 42.5                & 32.6                & \multicolumn{1}{c|}{22.0}                                        & 32.7                \\
\multicolumn{1}{l|}{}                                                                                                                    & \multicolumn{1}{l|}{}                                              & \multicolumn{1}{l|}{LVLM-text}                         & 29.9                & 19.5                & 64.1                & 48.7                & 37.5                & \multicolumn{1}{c|}{20.3}                                        & 36.7                \\
\multicolumn{1}{l|}{\multirow{-3}{*}{\begin{tabular}[c]{@{}l@{}}LLaVA-1.5\cite{liu2023improved}\\ (Vicuna-v1.5-7B\cite{chiang2023vicuna})\end{tabular}}}                              & \multicolumn{1}{l|}{\multirow{-3}{*}{7B}}                          & \multicolumn{1}{l|}{LVLM}                              & 34.4                & 65.0                & 68.7                & 55.6                & 65.6                & \multicolumn{1}{c|}{23.6}                                        & 52.2                \\
\rowcolor[HTML]{E4E4E4} 
\multicolumn{1}{l|}{\cellcolor[HTML]{E4E4E4}}                                                                                            & \multicolumn{1}{l|}{\cellcolor[HTML]{E4E4E4}}                      & \multicolumn{1}{l|}{\cellcolor[HTML]{E4E4E4}LLM}       & 32.8                & 8.9                 & 64.0                & 48.3                & 31.9                & \multicolumn{1}{c|}{\cellcolor[HTML]{E4E4E4}18.9}                & 34.1                \\
\rowcolor[HTML]{E4E4E4} 
\multicolumn{1}{l|}{\cellcolor[HTML]{E4E4E4}}                                                                                            & \multicolumn{1}{l|}{\cellcolor[HTML]{E4E4E4}}                      & \multicolumn{1}{l|}{\cellcolor[HTML]{E4E4E4}LVLM-text} & 34.2                & {\ul \textbf{26.2}} & {\ul \textbf{71.9}} & 63.3                & 38.1                & \multicolumn{1}{c|}{\cellcolor[HTML]{E4E4E4}{\ul \textbf{29.4}}} & 43.9 \\
\rowcolor[HTML]{E4E4E4} 
\multicolumn{1}{l|}{\multirow{-3}{*}{\cellcolor[HTML]{E4E4E4}\begin{tabular}[c]{@{}l@{}}InternLM2-XC2\cite{dong2024internlm}\\ (InternLM2-7B\cite{team2023internlm})\end{tabular}}}    & \multicolumn{1}{l|}{\multirow{-3}{*}{\cellcolor[HTML]{E4E4E4}7B}}  & \multicolumn{1}{l|}{\cellcolor[HTML]{E4E4E4}LVLM}      & 41.7                & 79.6                & 96.7                & 81.4                & 74.9                & \multicolumn{1}{c|}{\cellcolor[HTML]{E4E4E4}57.4}                & 72.0                \\
\multicolumn{1}{l|}{}                                                                                                                    & \multicolumn{1}{l|}{}                                              & \multicolumn{1}{l|}{LLM}                               & 19.8                & 8.4                 & 52.7                & 42.6                & 7.6                 & \multicolumn{1}{c|}{20.5}                                        & 25.3                \\
\multicolumn{1}{l|}{}                                                                                                                    & \multicolumn{1}{l|}{}                                              & \multicolumn{1}{l|}{LVLM-text}                         & 32.4                & 15.6 & 71.1 & 56.8                & 36.1                & \multicolumn{1}{c|}{25.0}                         & 39.5 \\
\multicolumn{1}{l|}{\multirow{-3}{*}{\begin{tabular}[c]{@{}l@{}}Monkey-Chat\cite{li2023monkey}\\ (Qwen-7B\cite{bai2023qwen})\end{tabular}}}                                   & \multicolumn{1}{l|}{\multirow{-3}{*}{10B}}                         & \multicolumn{1}{l|}{LVLM}                              & 37.1                & 71.0                & 82.4                & 68.5                & 69.1                & \multicolumn{1}{c|}{34.0}                                        & 60.4                \\
\rowcolor[HTML]{E4E4E4} 
\multicolumn{1}{l|}{\cellcolor[HTML]{E4E4E4}}                                                                                            & \multicolumn{1}{l|}{\cellcolor[HTML]{E4E4E4}}                      & \multicolumn{1}{l|}{\cellcolor[HTML]{E4E4E4}LLM}       & 29.9                & 10.3                & 58.9                & 42.5                & 32.6                & \multicolumn{1}{c|}{\cellcolor[HTML]{E4E4E4}22.0}                & 32.7                \\
\rowcolor[HTML]{E4E4E4} 
\multicolumn{1}{l|}{\cellcolor[HTML]{E4E4E4}}                                                                                            & \multicolumn{1}{l|}{\cellcolor[HTML]{E4E4E4}}                      & \multicolumn{1}{l|}{\cellcolor[HTML]{E4E4E4}LVLM-text} & 30.1                & 15.5                & 54.6                & 52.5                & 36.7                & \multicolumn{1}{c|}{\cellcolor[HTML]{E4E4E4}25.0}                & 35.7                \\
\rowcolor[HTML]{E4E4E4} 
\multicolumn{1}{l|}{\multirow{-3}{*}{\cellcolor[HTML]{E4E4E4}\begin{tabular}[c]{@{}l@{}}CogVLM-Chat\cite{wang2023cogvlm}\\ (Vicuna-v1.5-7B\cite{chiang2023vicuna})\end{tabular}}}    & \multicolumn{1}{l|}{\multirow{-3}{*}{\cellcolor[HTML]{E4E4E4}17B}} & \multicolumn{1}{l|}{\cellcolor[HTML]{E4E4E4}LVLM}      & 34.2                & 63.4                & 66.3                & 63.3                & 68.7                & \multicolumn{1}{c|}{\cellcolor[HTML]{E4E4E4}34.7}                & 55.1                \\
\multicolumn{1}{l|}{}                                                                                                                    & \multicolumn{1}{l|}{}                                              & \multicolumn{1}{l|}{LLM}                               & 37.1                & 10.5                & 53.6                & 57.3                & 37.3                & \multicolumn{1}{c|}{21.7}                                        & 36.3                \\
\multicolumn{1}{l|}{}                                                                                                                    & \multicolumn{1}{l|}{}                                              & \multicolumn{1}{l|}{LVLM-text}                         & 37.3                & 23.2                & 68.6                & 59.9                & {\ul \textbf{41.0}} & \multicolumn{1}{c|}{22.7}                                        & 42.1                \\
\multicolumn{1}{l|}{\multirow{-3}{*}{\begin{tabular}[c]{@{}l@{}}Yi-VL\cite{young2024yi}\\ (Yi-34B\cite{young2024yi})\end{tabular}}}                                          & \multicolumn{1}{l|}{\multirow{-3}{*}{34B}}                         & \multicolumn{1}{l|}{LVLM}                              & 43.2                & 71.5                & 75.3                & 65.9                & 68.1                & \multicolumn{1}{c|}{25.6}                                        & 58.3                \\
\rowcolor[HTML]{E4E4E4} 
\multicolumn{1}{l|}{\cellcolor[HTML]{E4E4E4}}                                                                                            & \multicolumn{1}{l|}{\cellcolor[HTML]{E4E4E4}}                      & \multicolumn{1}{l|}{\cellcolor[HTML]{E4E4E4}LLM}       & 37.6                & 20.1                & 69.4                & 60.2                & 35.0                & \multicolumn{1}{c|}{\cellcolor[HTML]{E4E4E4}17.9}                & 40.0                \\
\rowcolor[HTML]{E4E4E4} 
\multicolumn{1}{l|}{\cellcolor[HTML]{E4E4E4}}                                                                                            & \multicolumn{1}{l|}{\cellcolor[HTML]{E4E4E4}}                      & \multicolumn{1}{l|}{\cellcolor[HTML]{E4E4E4}LVLM-text} & 41.7                & 23.9                & 70.3                & {\ul \textbf{65.0}} & 40.5                & \multicolumn{1}{c|}{\cellcolor[HTML]{E4E4E4}24.0}                & {\ul\textbf{44.2}}               \\
\rowcolor[HTML]{E4E4E4} 
\multicolumn{1}{l|}{\multirow{-3}{*}{\cellcolor[HTML]{E4E4E4}\begin{tabular}[c]{@{}l@{}}InternVL-Chat-v1.2\cite{chen2023internvl}\\ (NH2-Yi-34B\cite{nousyi34b})\end{tabular}}} & \multicolumn{1}{l|}{\multirow{-3}{*}{\cellcolor[HTML]{E4E4E4}40B}} & \multicolumn{1}{l|}{\cellcolor[HTML]{E4E4E4}LVLM}      & 49.1                & 82.4                & 82.5                & 78.5                & 75.4                & \multicolumn{1}{c|}{\cellcolor[HTML]{E4E4E4}47.7}                & 69.3                \\
\multicolumn{1}{l|}{}                                                                                                                    & \multicolumn{1}{l|}{}                                              & \multicolumn{1}{l|}{LLM}                               & 25.7                & 8.6                 & 57.2                & 48.7                & 13.5                & \multicolumn{1}{c|}{23.4}                                        & 29.5                \\
\multicolumn{1}{l|}{}                                                                                                                    & \multicolumn{1}{l|}{}                                              & \multicolumn{1}{l|}{LVLM-text}                         & {\ul \textbf{43.6}} & 20.5                & 68.4                & 61.1                & 39.9                & \multicolumn{1}{c|}{28.4}                                        & 43.7                \\
\multicolumn{1}{l|}{\multirow{-3}{*}{\begin{tabular}[c]{@{}l@{}}Sphinx-X-MoE\cite{gao2024sphinx}\\ (Mixtral-8x7B\cite{jiang2024mixtral})\end{tabular}}}                             & \multicolumn{1}{l|}{\multirow{-3}{*}{57B}}                         & \multicolumn{1}{l|}{LVLM}                              & 44.8                & 69.2                & 72.2                & 65.0                & 71.1                & \multicolumn{1}{c|}{38.1}                                        & 60.1                \\ \bottomrule
\end{tabular}
}
\end{table*}

\begin{figure*}[t]
    \centering
    \includegraphics[width=\linewidth]{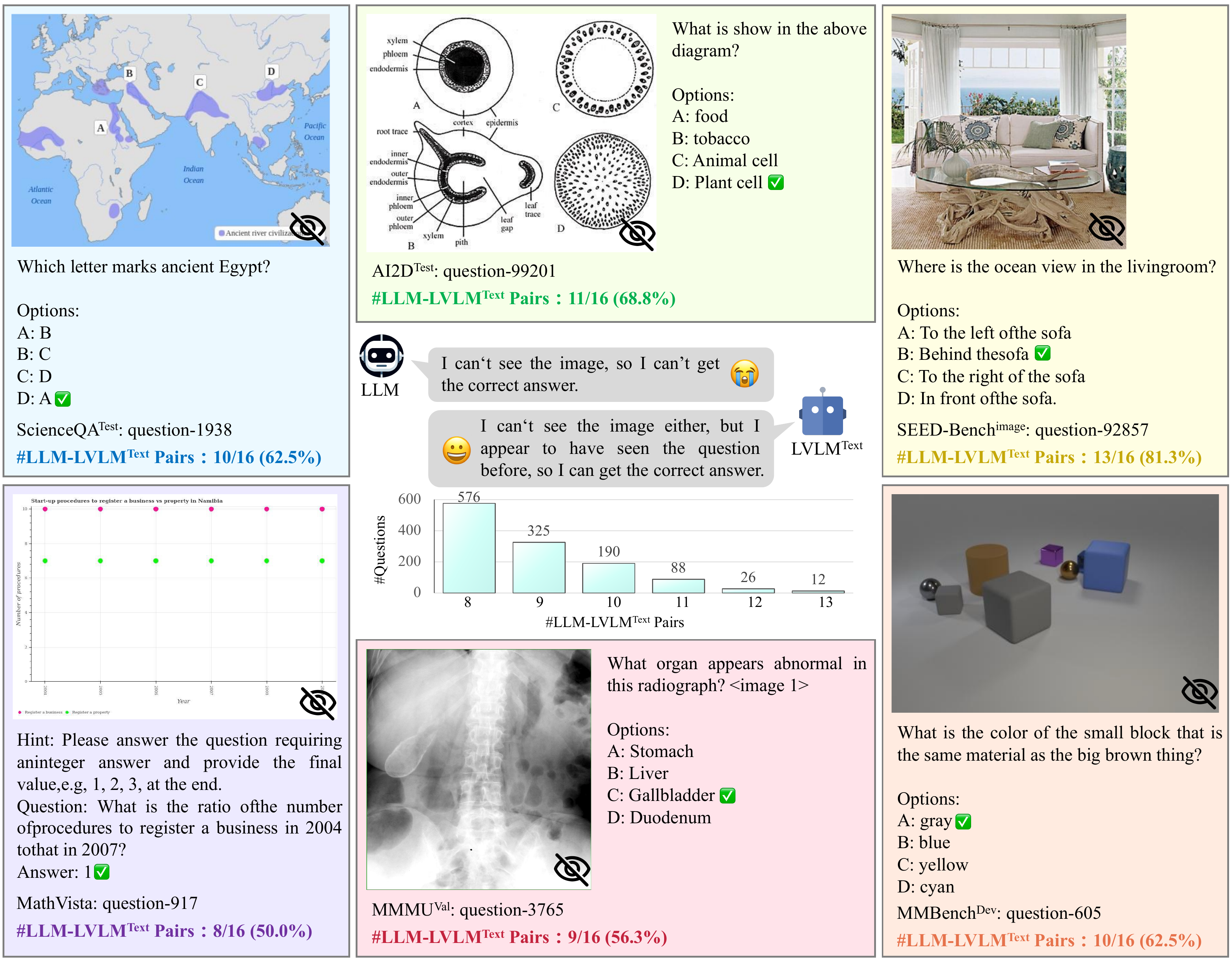}
    \captionsetup{font={footnotesize}}
    \caption{\textbf{Illustration of data leakage during LVLMs' multi-modal training processes.} We showcase samples that LLMs cannot answer correctly but LVLMs without accessing images (LVLM-text) can. Each LLM-LVLM$^{text}$ pair represents an LLM and its corresponding LVLM without accessing images, totaling 16 pairs. The chart in the center tallies the number of samples in existing benchmarks hit by more than half of the LLM-LVLM$^{text}$ pairs, underscoring the issue of data leakage during the multi-modal training process.}
    \label{fig:comp_llm_lvlm-text}
\end{figure*}

\textbf{Second issue: unintentional data leaking exists in LLM and LVLM training.} Although the community has the trend towards developing new multi-modal benchmarks to assess LVLMs' capabilities from various dimensions, there is scant consideration for fairness and reliability during evaluation. Training LLMs and LVLMs requires vast and diverse data, inevitably leading to the leakage of evaluation samples. Such incidents are usually unintended, as it's impractical to predict which data will be used in future evaluation benchmarks during the preparation for training corpus.

Figure \ref{fig:llm_case} (c) showcases an evaluation sample leaked by LLMs. Though the question requires an understanding of image content, 16 out of 22 tested LLMs astonishingly provide the correct response by "recalling" their training data. To quantitatively support our observations, we evaluate 22 leading LLMs across 6 popular benchmarks and report the 0-shot results in Table \ref{tab:llm} and 2-shot results in Table \ref{tab:llm_2_shot}. Specifically, we find the 2-shot evaluation strategy is more stable than the 0-shot to reduce refusal for answering and align answer formats. Under the impact of vision-independent samples and data leakage from LLMs, GeminiPro \cite{team2023gemini} and Qwen1.5-72B \cite{bai2023qwen} achieve a remarkable average accuracy of 41.4\% and 43.6\% under the 2-shot setting, outperforming random choice by 20.4\% and 22.6\%, respectively. Furthermore, Qwen1.5-72B achieves a score of 42.4\% on MMMU \cite{yue2023mmmu}, even surpassing the performance of the majority of LVLMs with accessing images. This result serves as a reminder: if we only consider the final accuracy on benchmarks when evaluating LVLMs, potential data leakage from LLMs could lead to unfair comparisons.

In Figure \ref{fig:llm_case} (d) and Figure \ref{fig:comp_llm_lvlm-text}, we showcase some examples where original LLMs fail, but LVLMs without accessing images succeed. Despite these questions requiring image content for accurate answers, the LVLMs without accessing images are capable of correctly answering these questions which stump original LLMs. To further support our hypotheses of data leakage during LVLMs' multi-modal training, we conduct an intriguing experiment. We remove the image inputs for LVLMs and only utilize textual questions and options for evaluation, with the results reported in Table \ref{tab:lvlm}. We compare the performance gains of LVLMs set to receive only text inputs (LVLM-text) against their corresponding LLM bases (LLM) to quantitatively assess the degree of data leakage in LVLMs' multi-modal training. As shown in Table \ref{tab:lvlm}, most LVLMs exhibit varying degrees of data leakage during multi-modal training. For example, the LLMs of Sphinx-X-8x7B \cite{gao2024sphinx} and Monkey-Chat \cite{li2023monkey}, show a respective average performance gain of 14.1\% and 14.2\% compared to their original LLMs.

Drawing from our observations, we posit that the issue of data leakage in multi-modal datasets is a significant concern that warrants attention. For research to be transparent and equitable, it is imperative to account for and mitigate such leakage. This will ensure that the performance of models is evaluated on their true ability to integrate and interpret multi-modal data, rather than their capacity to memorize specific samples. A proper benchmark would be a crucial step toward advancing the field of multi-modal language model research.

\section{MMStar}
\label{sec:mmstar}
With the aforementioned analysis, we present an elite vision-dependent multi-modal benchmark, dubbed as \textbf{MMStar}. In Section \ref{sec:data_curation}, we elaborate on the data curation process of MMStar. Section \ref{sec:core_capability} provides a detailed analysis of the constructed MMStar benchmark. In Section \ref{sec:mg_ml}, we introduce two benchmark-specific metrics developed to evaluate the degree of data leakage as well as the actual performance gains in multimodal capabilities from the multi-modal training.
\subsection{Data Curation Process}
\label{sec:data_curation}
\textbf{Criteria for data curation.} The evaluation samples for constructing the MMStar benchmark should meet three fundamental criteria: 1) \textbf{Visual dependency.} The collected samples can be correctly answered only based on understanding the visual content; 2) \textbf{Minimal data leakage.} The collected samples should minimize the risk of unintentional inclusion in LLMs' training corpus, or be effectively transformed from uni-modal to multi-modal formats to prevent LLMs from "recalling" the correct answers; 3) \textbf{Requiring advanced multi-modal capabilities for resolution.} In addition to ensuring fairness and reliability by adhering to the above criteria, we also aim for samples to cover various difficulty levels. We expect to comprehensively capture LVLMs' multi-modal capabilities with succinct high-quality samples. 

\begin{figure*}[t]
    \centering
    \includegraphics[width=\textwidth]{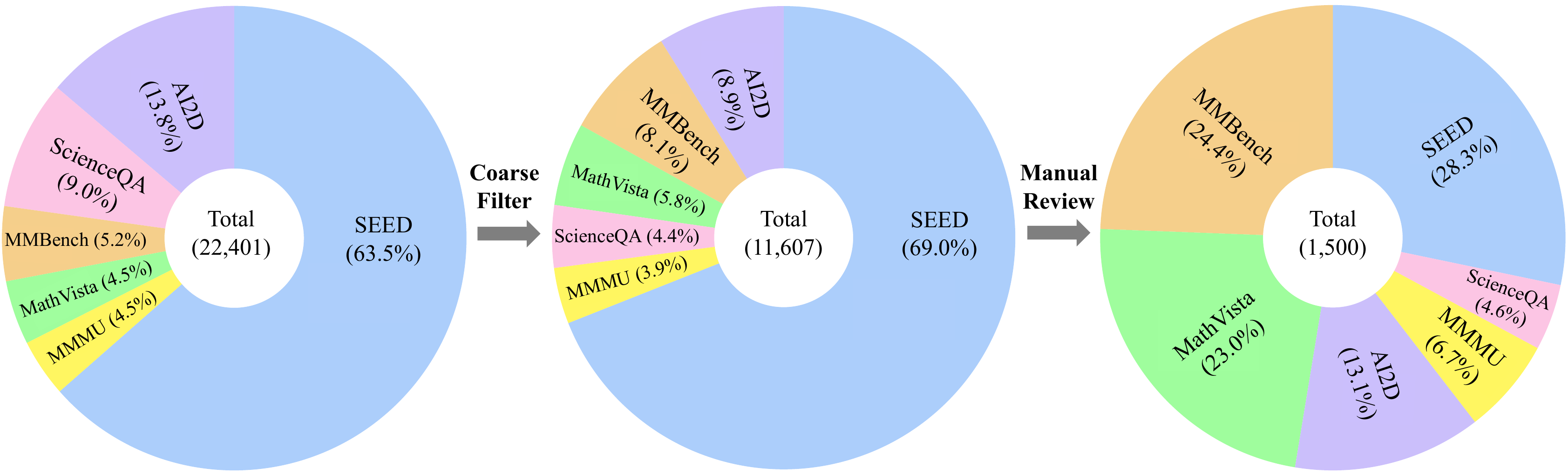}
    \captionsetup{font={footnotesize}}
    \caption{\textbf{Statics of the data sources during the data curation process.} After applying the coarse filter process and manual review, we narrow down from a total of 22,401 samples to 11,607 candidate samples and finally select 1,500 high-quality samples to construct our MMStar benchmark.}
    \label{fig:data_source}
\end{figure*}

\textbf{Data filter.} We first choose two benchmarks \cite{liu2023mmbench,li2023seed} focused on natural images and four centered on scientific and technical knowledge \cite{yue2023mmmu,lu2022learn,Kembhavi2016ADI,lu2023mathvista} for our sample collection. We then develop an automated pipeline to preliminarily filter out samples that do not meet the first two criteria. Specifically, we employ two closed-source LLMs \cite{team2023gemini,chatgpt} and six open-source LLMs \cite{bai2023qwen,team2023internlm,young2024yi,bi2024deepseek,jiang2024mixtral,touvron2023llama} sizing 20B or larger to serve as inspectors. These open-source LLMs are applied with a 2-shot in-context inference strategy to minimize response refusals and ensure consistency in answer formatting. Following this, we evaluate the sample pool with these LLM inspectors, documenting the hit frequency for each evaluation sample. Finally, we only retain those samples with hit counts of two or fewer hits, indicating that around 75\%  of LLM inspectors fail to provide the correct answer. As illustrated in Figure \ref{fig:data_source}, following this initial coarse filtering, our sample pool was reduced from 22,401 to 11,607.

\textbf{Manual review.} After the coarse filtering with LLM inspectors, we further employ three experts to conduct the manual review process to ensure: 1) each sample's answer should be based on the understanding of visual content; 2) selected samples should cover a comprehensive range of capability assessment dimensions; 3) most samples should require LVLMs to possess advanced multi-modal abilities for resolution. To expedite the manual selection of samples with varying difficulty levels for LVLMs, we tally the hit counts of all 16 LVLMs on the coarsely filtered samples and split them into four difficulty categories: easy (12-16), moderate (8-11), hard (4-7), and tough (0-3). Finally, after considering both the diversity of capability dimensions and difficulty levels, we manually curated \textbf{1,500} high-quality samples from the coarsely filtered set. Figure \ref{fig:data_source} showcases the detailed composition of data sources for our final selection of samples.

\subsection{Core Capabilities}
\label{sec:core_capability}
\begin{figure*}[t]
    \centering
    \includegraphics[width=0.7\textwidth]{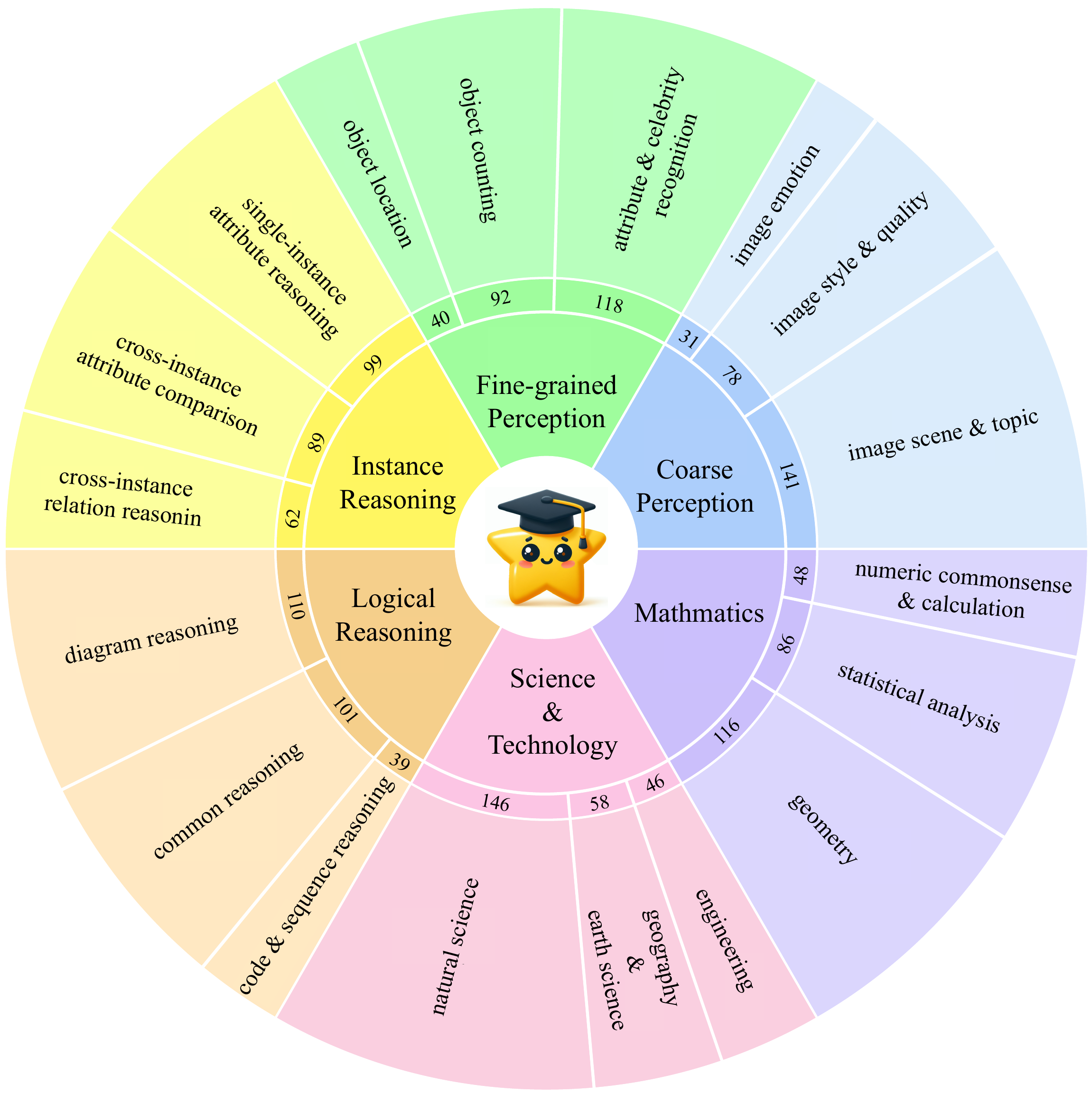}
    \captionsetup{font={footnotesize}}
    \caption{\textbf{Distribution of capability dimensions on the MMStar benchmark.} In MMStar, we display 6 core capabilities in the inner ring, with 18 detailed axes presented in the outer ring. The middle ring showcases the number of samples for each detailed dimension. Each core capability contains a meticulously balanced 250 samples. We further ensure a relatively even distribution across the 18 detailed axes.}
    \label{fig:mmstar_axes}
\end{figure*}

We select and consolidate the dimensions used for assessing LVLMs' multi-modal capabilities in existing benchmarks and identify six core capability dimensions along with eighteen detailed axes. In Figure \ref{fig:mmstar_axes}, we provide statistics for each core capability and their detailed axes on the MMStar benchmark.

\textbf{Coarse Perception (CP).} This core dimension refers to the capability to understand and interpret the overarching characteristics and themes of an image without delving into the finer details. It encompasses a broad, holistic view of the visual content, enabling the identification of: 1) image style \& quality; 2) image scene \& topic; and 3) image emotion.

\textbf{Fine-grained Perception (FP).} This core dimension represents a sophisticated level of image understanding that focuses on the detailed and nuanced aspects of visual content. It involves a deep dive into the specifics of images: 1) attribute \& celebrity recognition; 2) object location; and 3) object counting. This core dimension unveils the subtle intricacies that coarse perception might overlook.

\textbf{Instance Reasoning (IR).} It encapsulates a set of advanced cognitive capabilities focused on understanding and interpreting individual and collective object attributes and interrelations within an image. This process goes beyond mere recognition, delving into the analytical assessment of: 1) single-instance attribute reasoning; 2) cross-instance attribute comparison; and 3) cross-instance relation reasoning. It is a critical component for systems requiring a deep semantic understanding of visual content, enabling nuanced interaction with and response to complex visual content.

\textbf{Logical Reasoning (LR).} This core dimension encompasses a sophisticated framework of cognitive processes designed to interpret, deduce, and infer conclusions from visual content through a structured approach to logic and reasoning. This multi-faceted capability marries the intuitive understanding of visual content with the structured rigor of logical deduction, enabling: 1) diagram reasoning; 2) code \& sequence reasoning; and 3) common reasoning.

\textbf{Science \& Technology (ST).} It consists of a comprehensive framework for the application and integration of knowledge across a broad spectrum of science and technology. This domain combines the theoretical underpinnings and practical applications of various fields: 1) natural science; 2) engineering; and 3) geography \& earth science.

\textbf{Mathematics (MA).} Math is a foundational pillar of logical and analytical reasoning and encompasses a broad spectrum of capabilities essential for understanding, applying, and interpreting quantitative and spatial information. We primarily consider three aspects for evaluating LVLMs' logical thinking prowess: 1) numeric commonsense \& calculation; 2) geometry; and 3) statistical analysis.

\subsection{Multi-modal Gain/Leakage}
\label{sec:mg_ml}
Given our observation of the potential for inadvertent leakage of some evaluation samples during the multi-modal training process, the vanilla evaluation approach struggles to reveal LVLMs' actual performance gains derived from multi-modal training and fails to enable fair comparison with other competitors. Therefore, we propose two novel metrics to separately assess the degree of data leakage and actual performance gain from the multi-modal training process. 

To calculate the multi-modal gain (MG) metric for a given LVLM on a particular benchmark, we need to compute the scores of the same LVLM with and without visual inputs, separately denoted as $S_v$ and $S_{wv}$. Then the MG metric can be derived from the following formulation:
\begin{equation}
    MG=S_v-S_{wv}.
\end{equation}
To calculate the multi-modal leakage (ML) metric, we need to compute the extra score of the given LVLM's LLM base (without any multi-modal training), denoted as $S_t$. Then the ML metric is formulated as follows:
\begin{equation}
    ML=\text{max}(0, S_{wv}-S_t).
\end{equation}

\section{Experiments}
In this section, we conduct a systematic analysis of the proposed MMStar benchmark along with the MG/ML metrics. We detail the experimental setup in Section \ref{sec:experimental_setups}, study and analyze the performance of 16 leading LVLMs on MMStar in Section \ref{sec:result_mmstar}, and extensively investigate the MG/ML metrics of 16 LVLMs across 6 popular benchmarks and our MMStar in Section \ref{sec:result_mg_ml}.

\renewcommand{\arraystretch}{1.1}
\begin{table*}[!t]
    \centering
    \footnotesize
    \captionsetup{font={footnotesize}}
    \caption {
    \textbf{LLMs failed to solve problems in MMStar and performed close to random guessing, visual content is necessary to solve MMStar.}
    We evaluate various LLMs on MMStar with the 2-shot inference strategy. We report the results of 2 closed-source LLMs and 20 open-source LLMs with varying sizes and architectures. We report the detailed results of the CP (coarse perception), FP (fine-grained perception), IR(instance reasoning), LR (logical reasoning), ST (science \& technology), and MA (mathematics) core capabilities. The {\ul\textbf{best}} results are highlighted in {\ul \textbf{bold and underlined.}}}
    \label{tab:llm_mmstar}
    \scalebox{1.08}{
    \setlength{\tabcolsep}{11pt}
    \begin{tabular}{lccccccc}
    \toprule
    \multicolumn{1}{l|}{Model}           & CP   & FP   & IR   & LR   & ST   & \multicolumn{1}{c|}{MA}   & Avg. \\ \midrule
    \multicolumn{8}{c}{\textit{Baselines}}                                                                              \\ \midrule
    \multicolumn{1}{l|}{Random Choice}                        & 23.7 & 24.5 & 25.3 & 24.3 & 24.8 & \multicolumn{1}{c|}{25.1} & 24.6 \\ \midrule
    \multicolumn{8}{c}{\textit{Closed-source LLMs}}                                                                     \\ \midrule
    \multicolumn{1}{l|}{GPT4-Turbo\cite{chatgpt}}      & 2.4  & 4.0  & 9.6  & 18.0 & 13.6 & \multicolumn{1}{c|}{25.6} & 12.2 \\
    \multicolumn{1}{l|}{Gemini-Pro\cite{team2023gemini}}      & {\ul\textbf{16.8}} & {\ul\textbf{13.6}} & {\ul\textbf{20.4}} & {\ul\textbf{24.4}} & {\ul\textbf{19.6}} & \multicolumn{1}{c|}{{\ul\textbf{28.8}}} & {\ul\textbf{20.6}} \\ \midrule
    \multicolumn{8}{c}{\textit{Open-source LLMs}}                                                                       \\ \midrule
    \multicolumn{1}{l|}{Qwen1.5-1.8B\cite{bai2023qwen}}    & 28.4 & 28.4 & 25.6 & 23.2 & {\ul\textbf{23.2}} & \multicolumn{1}{c|}{29.6} & {\ul\textbf{26.4}} \\
    \multicolumn{1}{l|}{Phi2-2.7B\cite{phi2}}       & 11.2 & 11.2 & 15.2 & 10.8 & 11.6 & \multicolumn{1}{c|}{12.0} & 12.0 \\
    \multicolumn{1}{l|}{Yi-6B-Chat\cite{young2024yi}}      & 23.6 & 19.2 & 28.4 & 25.2 & 12.4 & \multicolumn{1}{c|}{29.6} & 23.1 \\
    \multicolumn{1}{l|}{LLaMA2-7B\cite{touvron2023llama}}       & 28.0 & {\ul\textbf{30.4}} & 26.0 & 18.0 & 18.8 & \multicolumn{1}{c|}{21.6} & 23.8 \\
    \multicolumn{1}{l|}{Qwen-7B\cite{bai2023qwen}}         & 11.6 & 5.6  & 12.8 & 5.6  & 7.2  & \multicolumn{1}{c|}{0.4}  & 7.2  \\
    \multicolumn{1}{l|}{Deepseek-7B\cite{bi2024deepseek}}     & 26.8 & 16.0 & 28.4 & 21.6 & {\ul\textbf{23.2}} & \multicolumn{1}{c|}{25.6} & 23.6 \\
    \multicolumn{1}{l|}{InternLM2-7B\cite{team2023internlm}}    & 22.0 & 14.8 & 22.0 & 21.6 & 15.2 & \multicolumn{1}{c|}{23.2} & 19.8 \\
    \multicolumn{1}{l|}{Qwen1.5-7B\cite{bai2023qwen}}      & 15.6 & 8.0  & 9.2  & 9.2  & 15.2 & \multicolumn{1}{c|}{9.2}  & 11.1 \\
    \multicolumn{1}{l|}{Vicuna-v1.5-7B\cite{chiang2023vicuna}}  & 22.0 & 27.6 & 29.6 & 26.4 & 18.0 & \multicolumn{1}{c|}{24.4} & 24.7 \\
    \multicolumn{1}{l|}{Baichuan2-7B\cite{yang2023baichuan}}    & 20.8 & 18.4 & 27.6 & 18.8 & 18.8 & \multicolumn{1}{c|}{21.2} & 20.9 \\
    \multicolumn{1}{l|}{Mistral-7B\cite{jiang2023mistral}}      & 20.0 & 23.6 & 24.4 & 23.6 & 20.0 & \multicolumn{1}{c|}{27.2} & 23.1 \\
    \multicolumn{1}{l|}{LLaMA2-13B\cite{touvron2023llama}}      & 23.6 & 23.6 & 28.0 & 21.2 & 16.4 & \multicolumn{1}{c|}{10.4} & 20.5 \\
    \multicolumn{1}{l|}{Vicuna-v1.5-13B\cite{chiang2023vicuna}} & {\ul\textbf{32.8}} & 24.0 & {\ul\textbf{28.8}} & 17.6 & 22.0 & \multicolumn{1}{c|}{14.4} & 23.3 \\
    \multicolumn{1}{l|}{Baichuan2-13B\cite{yang2023baichuan}}   & 26.4 & 18.0 & 28.0 & 20.4 & 21.2 & \multicolumn{1}{c|}{25.6} & 23.3 \\
    \multicolumn{1}{l|}{InternLM2-20B\cite{team2023internlm}}   & 18.2 & 17.8 & 22.6 & 23.8 & 17.8 & \multicolumn{1}{c|}{13.4} & 18.9 \\
    \multicolumn{1}{l|}{Yi-34B\cite{young2024yi}}          & 20.4 & 18.0 & 24.0 & 24.0 & 14.4 & \multicolumn{1}{c|}{{\ul\textbf{30.8}}} & 21.9 \\
    \multicolumn{1}{l|}{Mixtral-8x7B\cite{jiang2024mixtral}}    & 24.4 & 17.6 & 19.2 & {\ul\textbf{28.0}} & 16.0 & \multicolumn{1}{c|}{33.6} & 23.1 \\
    \multicolumn{1}{l|}{Deepseek-67B\cite{bi2024deepseek}}    & 29.2 & 22.4 & 18.4 & 26.0 & 20.4 & \multicolumn{1}{c|}{22.4} & 23.1 \\
    \multicolumn{1}{l|}{LLaMA2-70B\cite{touvron2023llama}}      & 22.4 & 20.0 & 19.6 & 14.4 & 7.2  & \multicolumn{1}{c|}{9.6}  & 15.5 \\
    \multicolumn{1}{l|}{Qwen1.5-72B\cite{bai2023qwen}}     & 21.6 & 16.0 & 21.2 & 14.0 & 17.2 & \multicolumn{1}{c|}{27.2} & 19.5 \\ \bottomrule
    \end{tabular}
    }
\end{table*}

\subsection{Experimental Setups}
\label{sec:experimental_setups}
\textbf{Evaluation models.} 1) \textbf{Baselines}: We utilize random choice and frequent choice strategies to serve as the baselines. The former randomly selects an option as the answer, while the latter selects the most frequent option within each benchmark dimension. 2) \textbf{Large Language Models}: We prepare two closed-source LLMs, GPT4 \cite{chatgpt} and GeminiPro \cite{team2023gemini}, and 20 popular open-source LLMs sizing from 1.8B to 72B for text-only evaluation, such as Qwen series \cite{bai2023qwen}, LLaMA2 series \cite{touvron2023llama}, Phi2 \cite{phi2}, Vicuna series \cite{chiang2023vicuna}, Deepseek series \cite{bi2024deepseek}, InternLM2 series \cite{team2023internlm}, Baichuan2 series \cite{yang2023baichuan}, Yi series \cite{young2024yi}, Mistral series \cite{jiang2023mistral,jiang2024mixtral}. Additionally, all the open-source LLMs we used are their Chat versions. and 3) \textbf{Large Vision-Language Models}: We prepare two closed-source LVLMs, GPT4V \cite{gpt4v} and GeminiPro-Vision \cite{team2023gemini}, and 14 popular open-source LVLMs sizing from 3B to 60B, such as TinyLLaVA-3B \cite{zhou2024tinyllava}, Yi-VL series \cite{young2024yi}, Qwen-VL-Chat \cite{bai2023qwenvl}, LLaVA-1.5 series \cite{liu2023improved}, ShareGPT4V-7B \cite{chen2023sharegpt4v}, Monkey-Chat \cite{li2023monkey}, LLaVA-Next \cite{liu2024llavanext}, Deepseek-VL-7B \cite{lu2024deepseek}, LLaVA-Next-34B \cite{liu2024llavanext}, CogVLM-Chat-17B \cite{wang2023cogvlm}, InternVL-Chat-v1.2 \cite{chen2023internvl}, Sphinx-X-8x7B \cite{gao2024sphinx}.

\textbf{Implementation details.} For evaluating LLMs on existing benchmarks, we employ both 0-shot and 2-shot strategies and will specify which is utilized when reporting results. For evaluating LLMs on MMStar, the 0-shot strategy yields poor scores, making comparisons difficult. Therefore, we exclusively utilize the 2-shot strategy to decrease the frequency of refusal to answer. Moreover,  All LVLMs are evaluated utilizing the 0-shot strategy across all benchmarks to ensure a fair comparison. 
When evaluating LVLMs under the `LVLM-text' setting (\textit{i.e.} answer without the image), most LVLMs work well by simply removing the image tokens from their default input tokens.
However, GeminiPro-Vision \cite{team2023gemini} and CogVLM-Chat \cite{wang2023cogvlm} require the replacement of the original images with pure grey images to bypass image content input and operate correctly. Given that all questions are ensured to be converted into a multiple-choice format, we develop some heuristic matching rules to calculate accuracy, avoiding the cumbersome process of re-invoking GPT4 for answer extraction. Moreover, all experiments in this study are conducted within the same codebase modified from VLMEvalKit \cite{2023opencompass}, and utilize NVIDIA A100 GPUs for non-API-based evaluation.

\renewcommand{\arraystretch}{1.1}
\begin{table*}[!t]
    \centering
    \footnotesize
    \captionsetup{font={footnotesize}}
    \caption {\textbf{Evaluation of various LVLMs on MMStar.} We report the results of 2 closed-source LLMs and 14 open-source LLMs with varying sizes and architectures. We report the detailed results of the CP (coarse perception), FP (fine-grained perception), IR(instance reasoning), LR (logical reasoning), ST (science \& technology), and MA (mathematics) core capabilities. The {\ul \textbf{best}} results are highlighted in {\ul \textbf{bold and underlined.}} The \textit{{\todo{\textbf{worst}}}} results of multi-modal gain (MG) and multi-modal leakage (ML) metrics are in \textit{{\todo{\textbf{italic red}}}}.}
    \label{tab:lvlm_mmstar}
    \scalebox{0.96}{
    \setlength{\tabcolsep}{3pt}
    \begin{tabular}{lllccccccccc}
    \toprule
    \multicolumn{1}{l|}{Model} & \multicolumn{1}{l|}{LLM}             & \multicolumn{1}{l|}{Param.} & CP                                      & FP                       & IR                                      & LR                                      & ST                                      & \multicolumn{1}{c|}{MA}                  & \multicolumn{1}{c|}{Avg.}                & MG$\uparrow$                                            & ML$\downarrow$                                            \\ \midrule
    \multicolumn{12}{c}{\textit{Baselines}}                                                                                                                                                                                                                                                                                                                                                                                                                                \\ \midrule
    \multicolumn{1}{l|}{Random Choice} & \multicolumn{1}{l|}{-}      & \multicolumn{1}{l|}{-}     & 23.7                                    & 24.5                     & 25.3                                    & 24.3                                    & 24.8                                    & \multicolumn{1}{c|}{25.1}                & \multicolumn{1}{l|}{24.6}                & -                                             & -                                             \\ \midrule
    \multicolumn{12}{c}{\textit{Closed-source LVLMs}}                                                                                                                                                                                                                                                                                                                                                                                                                      \\ \midrule
    \multicolumn{1}{l|}{GeminiPro-Vision\cite{team2023gemini}} &\multicolumn{1}{l|}{GeminiPro\cite{team2023gemini}}   & \multicolumn{1}{l|}{-}     & 51.6                                    & 28.8                     & 50.8                                    & 46.0                                    & 28.4                                    & \multicolumn{1}{c|}{{\ul \textbf{50.0}}} & \multicolumn{1}{c|}{42.6}                & 27.4                                          & {\ul \textbf{0.0}}                            \\
    \multicolumn{1}{l|}{GPT4V (low)\cite{gpt4v}} &\multicolumn{1}{l|}{GPT4-Turbo\cite{chatgpt}}        & \multicolumn{1}{l|}{-}     & \multicolumn{1}{l}{62.0}                & 32.8                     & 55.2                                    & 48.0                                    & 33.6                                    & \multicolumn{1}{c|}{44.8}                & \multicolumn{1}{c|}{46.1}                & 32.6                                          & 1.3                                           \\
    \multicolumn{1}{l|}{GPT4V (high)\cite{gpt4v}} &\multicolumn{1}{l|}{GPT4-Turbo\cite{chatgpt}}       & \multicolumn{1}{l|}{-}     & {\ul \textbf{76.6}}                     & {\ul \textbf{51.4}}      & {\ul \textbf{66.6}}                     & {\ul \textbf{55.8}}                     & {\ul \textbf{42.6}}                     & \multicolumn{1}{c|}{49.8}                & \multicolumn{1}{c|}{{\ul \textbf{57.1}}} & {\ul \textbf{43.6}}                           & 1.3                                           \\ \midrule
    \multicolumn{12}{c}{\textit{Open-source LVLMs}}                                                                                                                                                                                                                                                                                                                                                                                                                        \\ \midrule
    \multicolumn{1}{l|}{TinyLLaVA\cite{zhou2024tinyllava}} &\multicolumn{1}{l|}{Phi2-2.7B\cite{phi2}}         & \multicolumn{1}{l|}{3B}    & 60.4                                    & 31.6                     & 50.8                                    & 30.4                                    & 18.0                                    & \multicolumn{1}{c|}{24.8}                & \multicolumn{1}{c|}{36.0}                & 16.4                                          & 7.6                                           \\
    \multicolumn{1}{l|}{Yi-VL\cite{young2024yi}}  &\multicolumn{1}{l|}{Yi-6B\cite{young2024yi}}            & \multicolumn{1}{l|}{6B}    & 58.0                                    & 33.6                     & 46.4                                    & 34.8                                    & 20.4                                    & \multicolumn{1}{c|}{34.0}                & \multicolumn{1}{c|}{37.9}                & 15.6                                          & {\ul \textbf{0.0}}                            \\
    \multicolumn{1}{l|}{LLaVA-1.5\cite{liu2023improved}} &\multicolumn{1}{l|}{Vicuna-v1.5-7B\cite{chiang2023vicuna}}          & \multicolumn{1}{l|}{7B}    & 58.8                                    & 24.0                     & 38.8                                    & 24.0                                    & 13.6                                    & \multicolumn{1}{c|}{22.8}                & \multicolumn{1}{c|}{30.3}                & {\color[HTML]{FE0000} \textit{\textbf{10.7}}} & {\ul \textbf{0.0}}                            \\
    \multicolumn{1}{l|}{ShareGPT4V\cite{chen2023sharegpt4v}} &\multicolumn{1}{l|}{Vicuna-v1.5-7B\cite{chiang2023vicuna}}         & \multicolumn{1}{l|}{7B}    & 58.8                                    & 28.0                     & 45.6                                    & 24.4                                    & 17.2                                    & \multicolumn{1}{c|}{24.0}                & \multicolumn{1}{c|}{33.0}                & 11.9                                          & {\ul \textbf{0.0}}                            \\
    \multicolumn{1}{l|}{InternLM-XC2\cite{dong2024internlm}} &\multicolumn{1}{l|}{InternLM2-7B\cite{team2023internlm}}         & \multicolumn{1}{l|}{7B}    & \multicolumn{1}{l}{{\ul \textbf{70.8}}} & \multicolumn{1}{l}{48.8} & \multicolumn{1}{l}{{\ul \textbf{65.2}}} & \multicolumn{1}{l}{{\ul \textbf{56.4}}} & \multicolumn{1}{l}{{\ul \textbf{42.0}}} & \multicolumn{1}{l|}{49.2}                & \multicolumn{1}{l|}{{\ul \textbf{55.4}}} & 28.1                                          & 7.5                                           \\
    \multicolumn{1}{l|}{Qwen-VL-Chat\cite{bai2023qwenvl}}&\multicolumn{1}{l|}{Qwen-7B\cite{bai2023qwen}}       & \multicolumn{1}{l|}{8B}    & 59.6                                    & 32.0                      & 50.8                                    & 29.2                                    & 22.0                                     & \multicolumn{1}{c|}{31.6}                & \multicolumn{1}{c|}{37.5}                & 23.9                                          & {\ul \textbf{0.0}}                            \\
    \multicolumn{1}{l|}{Deepseek-VL\cite{lu2024deepseek}} &\multicolumn{1}{l|}{Deepseek-7B\cite{bi2024deepseek}}        & \multicolumn{1}{l|}{8B}    & 64.0                                    & 30.8                     & 49.2                                    & 36.4                                    & 21.6                                    & \multicolumn{1}{c|}{20.4}                & \multicolumn{1}{c|}{37.1}                & 15.7                                          & {\ul \textbf{0.0}}                            \\
    \multicolumn{1}{l|}{Monkey-Chat\cite{li2023monkey}}&\multicolumn{1}{l|}{Qwen-7B\cite{bai2023qwen}}        & \multicolumn{1}{l|}{10B}   & 57.6                                    & 36.4                     & 51.6                                    & 33.2                                    & 26.4                                    & \multicolumn{1}{c|}{24.4}                & \multicolumn{1}{c|}{38.3}                & 13.5                                          & {\color[HTML]{FE0000} \textit{\textbf{17.6}}} \\
    
    \multicolumn{1}{l|}{LLaVA-1.5\cite{liu2023improved}} &\multicolumn{1}{l|}{Vicuna-v1.5-13B\cite{chiang2023vicuna}}          & \multicolumn{1}{l|}{13B}   & 58.8                                    & 28.0                     & 41.6                                    & 24.4                                    & 18.4                                    & \multicolumn{1}{c|}{25.6}                & \multicolumn{1}{c|}{32.8}                & 13.9                                          & {\ul \textbf{0.0}}                            \\
    \multicolumn{1}{l|}{CogVLM-Chat\cite{wang2023cogvlm}} &\multicolumn{1}{l|}{Vicuna-v1.5-7B\cite{chiang2023vicuna}}        & \multicolumn{1}{l|}{17B}   & 66.8                                    & 36.8                     & 49.2                                    & 31.2                                    & 23.6                                    & \multicolumn{1}{c|}{11.6}                & \multicolumn{1}{c|}{36.5}                & 14.9                                          & {\ul \textbf{0.0}}                            \\
    \multicolumn{1}{l|}{Yi-VL\cite{young2024yi}} &\multicolumn{1}{l|}{Yi-34B\cite{young2024yi}}              & \multicolumn{1}{l|}{34B}   & 53.2                                    & 31.2                     & 52.0                                    & 32.4                                    & 12.4                                    & \multicolumn{1}{c|}{35.2}                & \multicolumn{1}{c|}{36.1}                & 18.8                                          & {\ul \textbf{0.0}}                            \\
    \multicolumn{1}{l|}{LLaVA-Next\cite{liu2024llavanext}} &\multicolumn{1}{l|}{NH2-Yi-34B\cite{nousyi34b}}         & \multicolumn{1}{l|}{34B}   & 66.4                                    & {\ul \textbf{52.0}}      & 62.4                                    & 46.0                                    & 32.4                                    & \multicolumn{1}{c|}{{\ul \textbf{53.6}}} & \multicolumn{1}{c|}{52.1}                & 29.4                                          & 2.4                                           \\
    \multicolumn{1}{l|}{InternVL-Chat-V1.2\cite{chen2023internvl}} &\multicolumn{1}{l|}{NH2-Yi-34B\cite{nousyi34b}} & \multicolumn{1}{l|}{40B}   & 67.6                                    & 43.2                     & 61.2                                    & 47.2                                    & 24.0                                    & \multicolumn{1}{c|}{19.2}                & \multicolumn{1}{c|}{43.7}                & {\ul \textbf{32.6}}                           & {\ul \textbf{0.0}}                            \\
    \multicolumn{1}{l|}{Sphinx-X-MOE\cite{gao2024sphinx}} &\multicolumn{1}{l|}{Mixtral-8x7B\cite{jiang2024mixtral}}  & \multicolumn{1}{l|}{57B}   & 58.4                                    & 40.8                     & 47.6                                    & 35.2                                    & 19.2                                    & \multicolumn{1}{c|}{32.0}                & \multicolumn{1}{c|}{38.9}                & 14.8                                          & 1.0                                           \\ \bottomrule
    \end{tabular}
}
\end{table*}

\subsection{Results Analysis of MMStar}
\label{sec:result_mmstar}
In this section, we present a comprehensive comparison of various LLMs and LVLMs performed on our MMStar benchmark and summarize our key observations in the following parts.

\textbf{Observation from LLMs.} We comprehensively evaluate 2 closed-source LLMs and 20 open-source LLMs of varying sizes and architectures on the MMStar benchmark and report the results in Table \ref{tab:llm_mmstar}. Encouragingly, the performance of these LLMs is almost indistinguishable from random choice, effectively validating that the evaluation samples of our MMStar exhibit significant visual dependency and minimal data leakage from LLMs. Notably, the smallest model, Qwen1.5-1.8B, achieves the best score. We conjecture this is due to it suffering the least stringent safety restrictions, thereby reducing instances of refusal to answer. Moreover, among the six core capabilities of MMStar, science \& technology (ST) prove to be the most challenging dimension for LLMs. The best score on ST is only 23.2\%, significantly lower than the best scores of around 30\% in other dimensions. We speculate this may be that samples within the ST dimension have the least degree of data leakage from LLMs' training data.

\textbf{Observation from LVLMs.} We evaluate 2 closed-source and 14 open-source LVLMs on our MMStar, with the results reported in Table \ref{tab:lvlm_mmstar}. As shown in the table, GPT4V\cite{gpt4v} with a high-resolution setting can achieve the best average score of 57.1\% among all LVLMs. Increasing the resolution and number of image tokens can boost the average score from 46.1\% to 57.1\% for GPT4V, offering a positive signal to the research community. Among the open-source LVLMs, InternLM-Xcomposer2 \cite{dong2024internlm} achieves an impressive score of 55.4\%. LLaVA-Next \cite{liu2024llavanext} even surpasses GPT4V and GeminiPro-Vision \cite{team2023gemini} in the mathematics (MA) core capability. Notably, no LVLMs managed to reach a passing average score (\textit{i.e.} $60\%$) in the core capabilities of fine-grained perception (FP), logical reasoning (LR), science \& Technology (ST), and mathematics (MA), highlighting these dimensions as particularly challenging for existing LVLMs. Moreover, TinyLLaVA \cite{zhou2024tinyllava}, despite its modest 3B scale, outperformed some competitors of 7B and even 13B surprisingly, underscoring the potential of smaller-scale LVLMs. Additionally, even with the same architecture, ShareGPT4V-7B \cite{chen2023sharegpt4v} even outperforms LLaVA-1.5-13B with high-quality caption data. This result highlights the significance of high-quality caption data for LVLM performance to the community.

\renewcommand{\arraystretch}{1.15}
\begin{table*}[!t]
    \centering
    \footnotesize
    \captionsetup{font={footnotesize}}
    \caption {\textbf{Evaluation of various LVLMs on 7 Benchmarks with multi-modal gain (MG) and multi-modal leakage (ML) metrics.} We report the results of 2 closed-source LLMs and 14 open-source LLMs with varying sizes and architectures. The bottom row represents the average across models for the same benchmark, while the rightmost column shows the average across benchmarks for the same LVLM. The {\ul \textbf{best}} results are highlighted in {\ul \textbf{bold and underlined.}} The \textit{{\todo{\textbf{worst}}}} results of MG and ML metrics are in \textit{{\todo{\textbf{italic red}}}}.}
    \label{tab:mg_ml}
    \scalebox{0.85}{
    \setlength{\tabcolsep}{1.5pt}
\begin{tabular}{lccccccccccccccccc}
\toprule
\multicolumn{1}{l|}{}                        & \multicolumn{1}{c|}{}                         & \multicolumn{2}{c|}{MMMU}                                                                                         & \multicolumn{2}{c|}{MMB}                                                                                       & \multicolumn{2}{c|}{ScienceQA}                                                                                    & \multicolumn{2}{c|}{AI2D}                                                                                         & \multicolumn{2}{c|}{SEED}                                                                                          & \multicolumn{2}{c|}{MathVista}                                                                                    & \multicolumn{2}{c|}{MMStar}                                                                                        & \multicolumn{2}{c}{Avg.}                                                                      \\ \cline{3-18} 
\multicolumn{1}{l|}{\multirow{-2}{*}{Model}} & \multicolumn{1}{c|}{\multirow{-2}{*}{Param.}} & MG$\uparrow$                                           & \multicolumn{1}{c|}{ML$\downarrow$}                                            & MG$\uparrow$                                            & \multicolumn{1}{c|}{ML$\downarrow$}                                            & MG$\uparrow$                                           & \multicolumn{1}{c|}{ML$\downarrow$}                                            & MG$\uparrow$                                           & \multicolumn{1}{c|}{ML$\downarrow$}                                            & MG$\uparrow$                                            & \multicolumn{1}{c|}{ML$\downarrow$}                                            & MG$\uparrow$                                           & \multicolumn{1}{c|}{ML$\downarrow$}                                            & MG$\uparrow$                                            & \multicolumn{1}{c|}{ML$\downarrow$}                                            & MG$\uparrow$                                            & ML$\downarrow$                                            \\ \midrule
\multicolumn{18}{c}{\textit{Closed-source LVLMs}}                                                                                                                                                                                                                                                                                                                                                                                                                                                                                                                                                                                                                                                                                                                                                                                                                                                                                                                                                                                                    \\ \midrule
\multicolumn{1}{l|}{GPT4V\cite{gpt4v}}                   & \multicolumn{1}{c|}{-}                        & {\ul \textbf{8.5}}                           & \multicolumn{1}{c|}{3.9}                                           & {\ul \textbf{52.0}}                           & \multicolumn{1}{c|}{5.4}                                           & 13.2                                         & \multicolumn{1}{c|}{3.9}                                           & 12.8                                         & \multicolumn{1}{c|}{2.8}                                           & {\ul \textbf{43.2}}                           & \multicolumn{1}{c|}{18.3}                                          & {\ul \textbf{19.3}}                          & \multicolumn{1}{c|}{{\ul \textbf{1.2}}}                            & {\ul \textbf{32.6}}                           & \multicolumn{1}{c|}{1.3}                                           & {\ul \textbf{25.9}}                           & 5.3                                           \\
\multicolumn{1}{l|}{GeminiPro-Vision\cite{team2023gemini}}        & \multicolumn{1}{c|}{-}                        & 5.0                                          & \multicolumn{1}{c|}{{\ul \textbf{0.0}}}                            & 51.4                                          & \multicolumn{1}{c|}{{\ul \textbf{0.0}}}                            & {\ul \textbf{14.3}}                          & \multicolumn{1}{c|}{{\ul \textbf{0.0}}}                            & {\ul \textbf{13.5}}                          & \multicolumn{1}{c|}{{\ul \textbf{0.0}}}                            & 36.4                                          & \multicolumn{1}{c|}{{\ul \textbf{0.0}}}                            & 11.5                                         & \multicolumn{1}{c|}{{\ul \textbf{1.2}}}                            & 27.4                                          & \multicolumn{1}{c|}{{\ul \textbf{0.0}}}                            & 22.8                                          & {\ul \textbf{0.2}}                            \\ \midrule
\multicolumn{18}{c}{\textit{Open-source LVLMs}}                                                                                                                                                                                                                                                                                                                                                                                                                                                                                                                                                                                                                                                                                                                                                                                                                                                                                                                                                                                                      \\ \midrule
\multicolumn{1}{l|}{TinyLLaVA\cite{zhou2024tinyllava}}               & \multicolumn{1}{c|}{3B}                       & 6.0                                          & \multicolumn{1}{c|}{10.0}                                          & 45.9                                          & \multicolumn{1}{c|}{13.8}                                          & 6.8                                          & \multicolumn{1}{c|}{15.2}                                          & 10.5                                         & \multicolumn{1}{c|}{13.2}                                          & 32.9                                          & \multicolumn{1}{c|}{10.8}                                          & 5.4                                          & \multicolumn{1}{c|}{1.5}                                           & 16.4                                          & \multicolumn{1}{c|}{7.6}                                           & 17.7                                          & 10.3                                          \\
\multicolumn{1}{l|}{Yi-VL\cite{young2024yi}}                   & \multicolumn{1}{c|}{6B}                       & 5.3                                          & \multicolumn{1}{c|}{7.4}                                           & 45.6                                          & \multicolumn{1}{c|}{14.1}                                          & 5.1                                          & \multicolumn{1}{c|}{9.4}                                           & {\color[HTML]{FE0000} \textit{\textbf{3.9}}} & \multicolumn{1}{c|}{{\color[HTML]{FE0000} \textit{\textbf{16.6}}}} & 29.2                                          & \multicolumn{1}{c|}{10.9}                                          & 3.8                                          & \multicolumn{1}{c|}{3.0}                                           & 15.6                                          & \multicolumn{1}{c|}{{\ul \textbf{0.0}}}                            & 15.5                                          & 8.8                                           \\
\multicolumn{1}{l|}{LLaVA-1.5\cite{liu2023improved}}               & \multicolumn{1}{c|}{7B}                       & 4.5                                          & \multicolumn{1}{c|}{{\ul \textbf{0.0}}}                            & {\color[HTML]{FE0000} \textit{\textbf{45.5}}} & \multicolumn{1}{c|}{9.2}                                           & 4.6                                          & \multicolumn{1}{c|}{5.2}                                           & 6.9                                          & \multicolumn{1}{c|}{6.2}                                           & 28.1                                          & \multicolumn{1}{c|}{4.9}                                           & 3.3                                          & \multicolumn{1}{c|}{{\ul \textbf{0.0}}}                            & {\color[HTML]{FE0000} \textit{\textbf{10.7}}} & \multicolumn{1}{c|}{{\ul \textbf{0.0}}}                            & {\color[HTML]{FE0000} \textit{\textbf{14.8}}} & 3.6                                           \\
\multicolumn{1}{l|}{ShareGPT4V\cite{chen2023sharegpt4v}}              & \multicolumn{1}{c|}{7B}                       & 3.5                                          & \multicolumn{1}{c|}{1.8}                                           & 49.1                                          & \multicolumn{1}{c|}{10.1}                                          & 4.2                                          & \multicolumn{1}{c|}{6.3}                                           & 8.5                                          & \multicolumn{1}{c|}{6.9}                                           & 31.7                                          & \multicolumn{1}{c|}{5.1}                                           & 3.0                                          & \multicolumn{1}{c|}{0.7}                                           & 11.9                                          & \multicolumn{1}{c|}{{\ul \textbf{0.0}}}                            & 16.0                                          & 4.4                                           \\
\multicolumn{1}{l|}{InternLM-XC2\cite{dong2024internlm}}            & \multicolumn{1}{c|}{7B}                       & 7.5                                          & \multicolumn{1}{c|}{1.4}                                           & 53.4                                          & \multicolumn{1}{c|}{{\color[HTML]{FE0000} \textit{\textbf{17.3}}}} & {\ul \textbf{24.8}}                          & \multicolumn{1}{c|}{7.9}                                           & {\ul \textbf{18.1}}                          & \multicolumn{1}{c|}{15.0}                                          & 36.8                                          & \multicolumn{1}{c|}{6.2}                                           & {\ul \textbf{28.0}}                          & \multicolumn{1}{c|}{{\color[HTML]{FE0000} \textit{\textbf{10.5}}}} & 28.1                                          & \multicolumn{1}{c|}{7.5}                                           & {\ul \textbf{28.1}}                           & 9.4                                          \\
\multicolumn{1}{l|}{Qwen-VL-Chat\cite{bai2023qwenvl}}            & \multicolumn{1}{c|}{8B}                       & {\ul \textbf{10.0}}                          & \multicolumn{1}{c|}{4.2}                                           & 49.6                                          & \multicolumn{1}{c|}{{\ul \textbf{0.3}}}                            & 11.0                                         & \multicolumn{1}{c|}{4.0}                                           & 12.3                                         & \multicolumn{1}{c|}{6.4}                                           & {\ul \textbf{44.5}}                           & \multicolumn{1}{c|}{11.9}                                          & 11.4                                         & \multicolumn{1}{c|}{0.3}                                           & 23.9                                          & \multicolumn{1}{c|}{{\ul \textbf{0.0}}}                            & 23.2                                          & 3.9                                           \\
\multicolumn{1}{l|}{Deepseek-VL\cite{lu2024deepseek}}             & \multicolumn{1}{c|}{8B}                       & 3.2                                          & \multicolumn{1}{c|}{10.6}                                          & 49.6                                          & \multicolumn{1}{c|}{15.5}                                          & 14.3                                         & \multicolumn{1}{c|}{10.8}                                          & 11.6                                         & \multicolumn{1}{c|}{14.9}                                          & 33.7                                          & \multicolumn{1}{c|}{23.1}                                          & 11.4                                         & \multicolumn{1}{c|}{3.3}                                           & 15.7                                          & \multicolumn{1}{c|}{{\ul \textbf{0.0}}}                            & 19.9                                          & 11.2                                          \\
\multicolumn{1}{l|}{Monkey-Chat\cite{li2023monkey}}             & \multicolumn{1}{c|}{10B}                      & 4.7                                          & \multicolumn{1}{c|}{12.6}                                          & 55.4                                          & \multicolumn{1}{c|}{7.2}                                           & 11.3                                         & \multicolumn{1}{c|}{{\color[HTML]{FE0000} \textit{\textbf{18.4}}}} & 11.7                                         & \multicolumn{1}{c|}{14.2}                                          & 33.0                                          & \multicolumn{1}{c|}{{\color[HTML]{FE0000} \textit{\textbf{28.5}}}} & 9.0                                          & \multicolumn{1}{c|}{4.5}                                           & 13.5                                          & \multicolumn{1}{c|}{{\color[HTML]{FE0000} \textit{\textbf{11.1}}}} & 19.8                                          & {\color[HTML]{FE0000} \textit{\textbf{13.8}}} \\
\multicolumn{1}{l|}{LLaVA-1.5\cite{liu2023improved}}               & \multicolumn{1}{c|}{13B}                      & 9.6                                          & \multicolumn{1}{c|}{{\ul \textbf{0.0}}}                            & 47.2                                          & \multicolumn{1}{c|}{9.8}                                           & 5.7                                          & \multicolumn{1}{c|}{7.0}                                           & 8.6                                          & \multicolumn{1}{c|}{7.2}                                           & 31.1                                          & \multicolumn{1}{c|}{10.7}                                          & 5.3                                          & \multicolumn{1}{c|}{1.5}                                           & 13.9                                          & \multicolumn{1}{c|}{{\ul \textbf{0.0}}}                            & 17.3                                          & 5.2                                           \\
\multicolumn{1}{l|}{CogVLM-Chat\cite{wang2023cogvlm}}             & \multicolumn{1}{c|}{17B}                      & 4.1                                          & \multicolumn{1}{c|}{0.2}                                           & 47.9                                          & \multicolumn{1}{c|}{5.2}                                           & 11.7                                         & \multicolumn{1}{c|}{{\ul \textbf{0.0}}}                            & 10.8                                         & \multicolumn{1}{c|}{10.0}                                          & 32.0                                          & \multicolumn{1}{c|}{4.1}                                           & 9.7                                          & \multicolumn{1}{c|}{3.0}                                           & 14.9                                          & \multicolumn{1}{c|}{{\ul \textbf{0.0}}}                            & 18.7                                          & {\ul \textbf{3.2}}                            \\
\multicolumn{1}{l|}{Yi-VL\cite{young2024yi}}                   & \multicolumn{1}{c|}{34B}                      & 5.9                                          & \multicolumn{1}{c|}{0.2}                                           & 48.3                                          & \multicolumn{1}{c|}{12.7}                                          & 6.7                                          & \multicolumn{1}{c|}{15.0}                                          & 6.0                                          & \multicolumn{1}{c|}{{\ul \textbf{2.6}}}                            & {\color[HTML]{FE0000} \textit{\textbf{27.1}}} & \multicolumn{1}{c|}{{\ul \textbf{3.7}}}                            & {\color[HTML]{FE0000} \textit{\textbf{2.9}}} & \multicolumn{1}{c|}{1.0}                                           & 18.8                                          & \multicolumn{1}{c|}{{\ul \textbf{0.0}}}                            & 16.5                                          & 5.0                                           \\
\multicolumn{1}{l|}{LLaVA-Next\cite{liu2024llavanext}}              & \multicolumn{1}{c|}{34B}                      & 6.6                                          & \multicolumn{1}{c|}{2.8}                                           & 54.7                                          & \multicolumn{1}{c|}{4.8}                                           & 11.2                                         & \multicolumn{1}{c|}{1.5}                                           & 12.8                                         & \multicolumn{1}{c|}{5.6}                                           & 34.1                                          & \multicolumn{1}{c|}{6.7}                                           & 16.5                                         & \multicolumn{1}{c|}{4.3}                                           & 29.4                                          & \multicolumn{1}{c|}{2.4}                                           & 23.6                                          & 4.0                                           \\
\multicolumn{1}{l|}{InternVL-Chat-v1.2\cite{chen2023internvl}}      & \multicolumn{1}{c|}{40B}                      & 7.4                                          & \multicolumn{1}{c|}{4.1}                                           & {\ul \textbf{58.5}}                           & \multicolumn{1}{c|}{3.8}                                           & 12.2                                         & \multicolumn{1}{c|}{0.9}                                           & 13.5                                         & \multicolumn{1}{c|}{4.8}                                           & 34.9                                          & \multicolumn{1}{c|}{5.5}                                           & 23.7                                         & \multicolumn{1}{c|}{6.1}                                           & {\ul \textbf{32.6}}                           & \multicolumn{1}{c|}{{\ul \textbf{0.0}}}                            & 26.1                                          & 3.6                                           \\
\multicolumn{1}{l|}{Sphinx-X-MoE\cite{gao2024sphinx}}            & \multicolumn{1}{c|}{57B}                      & {\color[HTML]{FE0000} \textit{\textbf{1.2}}} & \multicolumn{1}{c|}{{\color[HTML]{FE0000} \textit{\textbf{17.9}}}} & 48.7                                          & \multicolumn{1}{c|}{11.9}                                          & {\color[HTML]{FE0000} \textit{\textbf{3.8}}} & \multicolumn{1}{c|}{11.2}                                          & {\color[HTML]{FE0000} \textit{\textbf{3.9}}} & \multicolumn{1}{c|}{12.4}                                          & 31.2                                          & \multicolumn{1}{c|}{26.4}                                          & 9.7                                          & \multicolumn{1}{c|}{5.0}                                           & 14.8                                          & \multicolumn{1}{c|}{1.0}                                           & 16.2                                          & 12.3                                          \\ \midrule
\multicolumn{1}{l|}{Avg. across models}                    & \multicolumn{1}{c|}{-}                        & {\color[HTML]{FE0000} \textit{\textbf{5.8}}} & \multicolumn{1}{c|}{4.9}                                           & {\ul \textbf{50.1}}                           & \multicolumn{1}{c|}{8.9}                                           & 10.0                                         & \multicolumn{1}{c|}{7.4}                                           & 10.3                                         & \multicolumn{1}{c|}{8.7}                                           & 33.7                                          & \multicolumn{1}{c|}{{\color[HTML]{FE0000} \textit{\textbf{11.1}}}} & 10.8                                         & \multicolumn{1}{c|}{3.0}                                           & 20.0                                          & \multicolumn{1}{c|}{{\ul \textbf{1.9}}}                            & -                                             & -                                             \\ \bottomrule
\end{tabular}
    }
\end{table*}

\subsection{Results Analysis of MG/ML}
\label{sec:result_mg_ml}
In this section, we present the results of our proposed multi-modal gain (MG) and multi-modal leakage (ML) metrics of 16 LVLMs with varying sizes and architectures on 6 popular benchmarks and our MMStar benchmark. We then detail our observations and analyses from both the model and benchmark perspectives.

\textbf{Analysis from the model perspective.} In Table \ref{tab:mg_ml}, we illustrate the MG/ML (Multi-modal Gain/Multi-modal Leakage) metrics for each LVLM across each benchmark and provide an average MG/ML metric across all benchmarks in the final column. For closed-source LVLMs, GPT4V demonstrates notable performance gains attributed to its multi-modal training, while GeminiPro-Vision shows lesser data leakage during multi-modal training. This suggests that GPT4V may have utilized a broader range of multi-modal training data compared to GeminiPro-Vision. Among the open-source LVLMs, InternLM-XComposer2 achieves the highest average multi-modal gain of 28.1 across all benchmarks, whereas LLaVA-1.5-7B records the lowest at 14.8. This outcome is reasonable given that LLaVA-1.5-7B employed the least amount of multi-modal training data among these open-source LVLMs. Despite LLaVA-1.5-7B having the lowest average multi-modal gain, it exhibits minimal multi-modal leakage. Additionally, models like Monkey-Chat, Spinx-X-MoE, and Deepspeek-VL display higher degrees of multi-modal leakage, highlighting the need for the community to consider this factor for fair comparisons.

\textbf{Analysis from the benchmark perspective.} In the final row of Table \ref{tab:mg_ml}, we list the average multi-modal gain and multi-modal leakage for existing LVLMs across all benchmarks for analysis. MMBench registers the highest average multi-modal gain at 50.1, indicating a significant overlap between the domains covered by existing LVLMs' training data and MMBench. Conversely, MMMU exhibits the lowest average multi-modal gain at 5.8, suggesting a lesser degree of overlap between the domains of existing LVLMs' training corpora and those included in MMMU. Additionally, MMStar, as expected, has the lowest degree of multi-modal leakage at 1.9. 
This provides a comprehensive and fair arena for comparing existing LVLMs. Moreover, we believe evaluating existing LVLMs to derive average ML metrics can also be helpful to the following works in examining newly developed multi-modal benchmarks.

\section{Conclusion}
In this work, we dig into current evaluation works for large vision-language models (LVLMs) and identify two primary issues: 1) visual content is unnecessary for many samples, and 2) unintentional data leakage exists in LLM and LVLM training. To address these issues, we develop an elite vision-dependent multi-modal benchmark named MMStar and propose two metrics to measure the data leakage and actual performance gain in LVLMs' multi-modal training. MMStar undergoes the manual review of each sample, covering 6 core capabilities and 18 detailed axes for an in-depth evaluation of LVLMs' multimodal capabilities. In our evaluation of 16 diverse LVLMs on MMStar, even the best model scores under 60 on average. We also analyze the MG and ML metrics across 6 multimodal benchmarks and MMStar, providing valuable insights for the community on gathering multimodal training data and crafting new benchmarks. In the future, we plan to expand MMStar into a larger, online test set and explore dynamic evaluation methods to maintain sample visual dependency and reduce accidental data leakage into LLM's and LVLM's training corpora.

\appendix
\section{Cases of Lacking Visual Dependency}
\begin{figure*}[ht]
    \centering
    \includegraphics[width=0.95\linewidth]{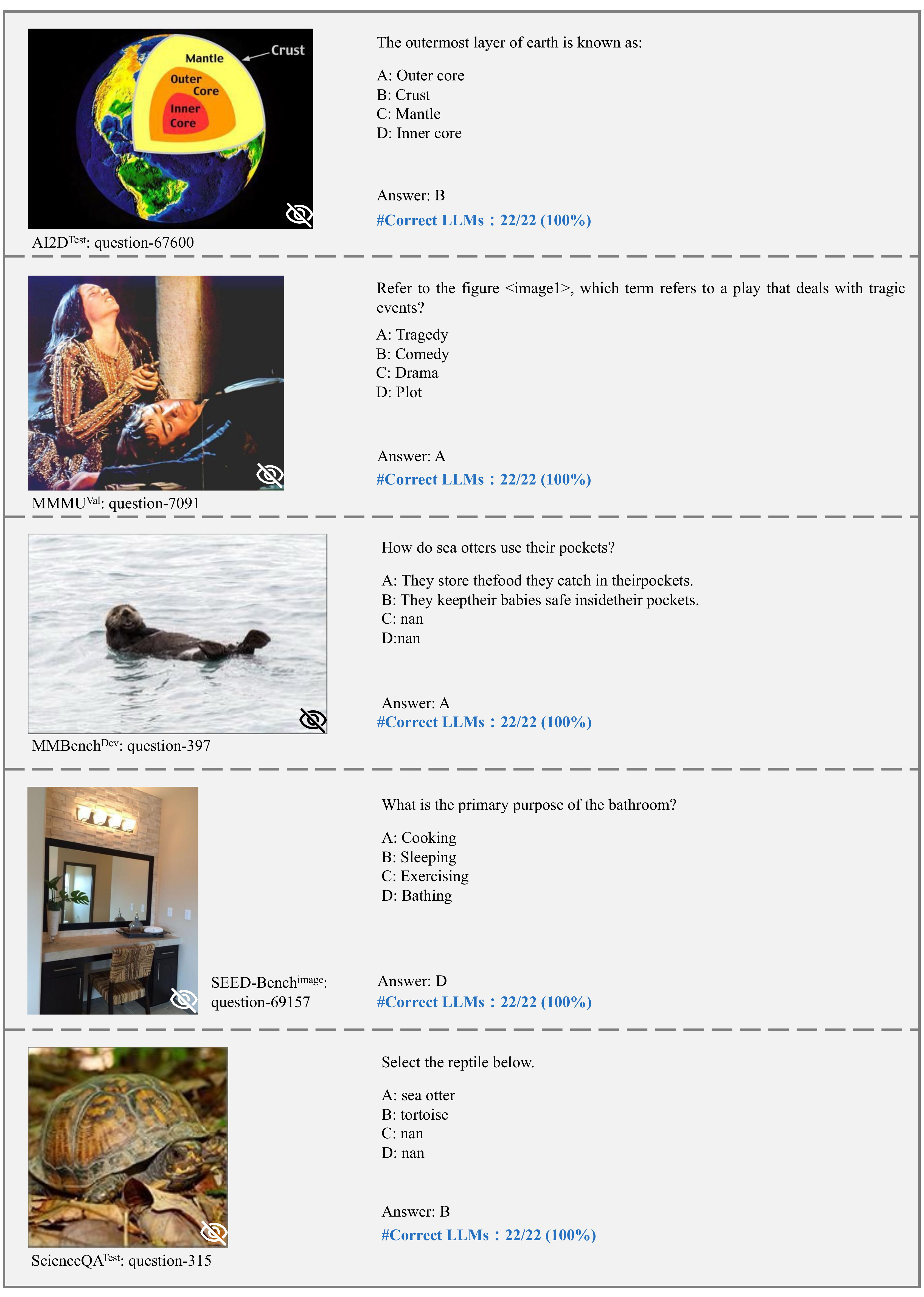}
    \captionsetup{font={footnotesize}}
    \caption{We highlight cases in existing benchmarks where evaluation samples lack the visual necessary.}
    \label{fig:llm_case1}
\end{figure*}
\clearpage

\section{Cases of Data Leakage in LLMs' Training Data}
\begin{figure*}[h]
    \centering
    \includegraphics[width=0.95\linewidth]{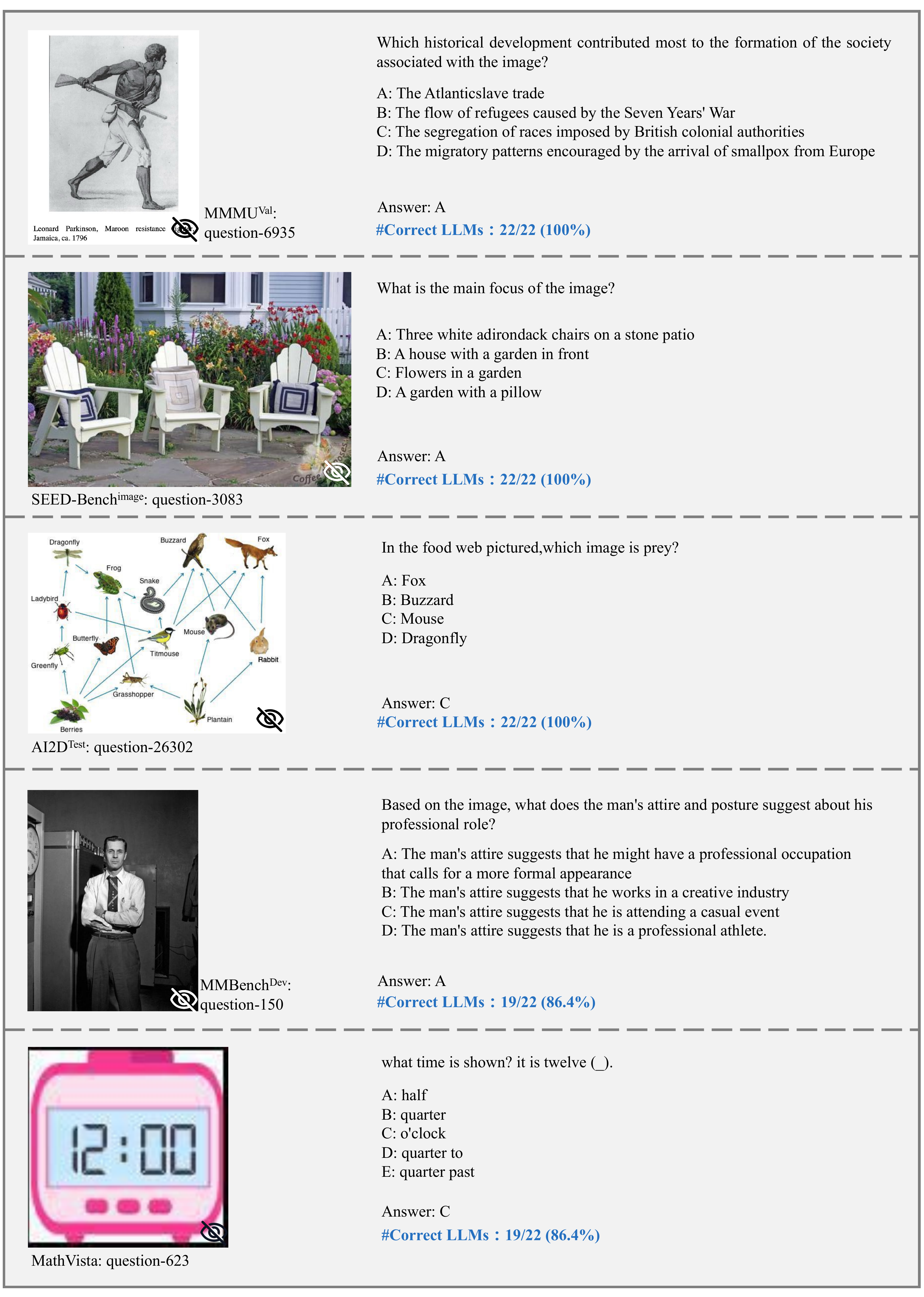}
    \captionsetup{font={footnotesize}}
    \caption{We highlight cases in existing benchmarks where evaluation samples are leaked into LLMs' training data.}
    \label{fig:llm_case2}
\end{figure*}
\clearpage

\section{Cases of Data Leakage in LVLMs' Multi-Modal Training Data}
\begin{figure*}[h]
    \centering
    \includegraphics[width=0.95\linewidth]{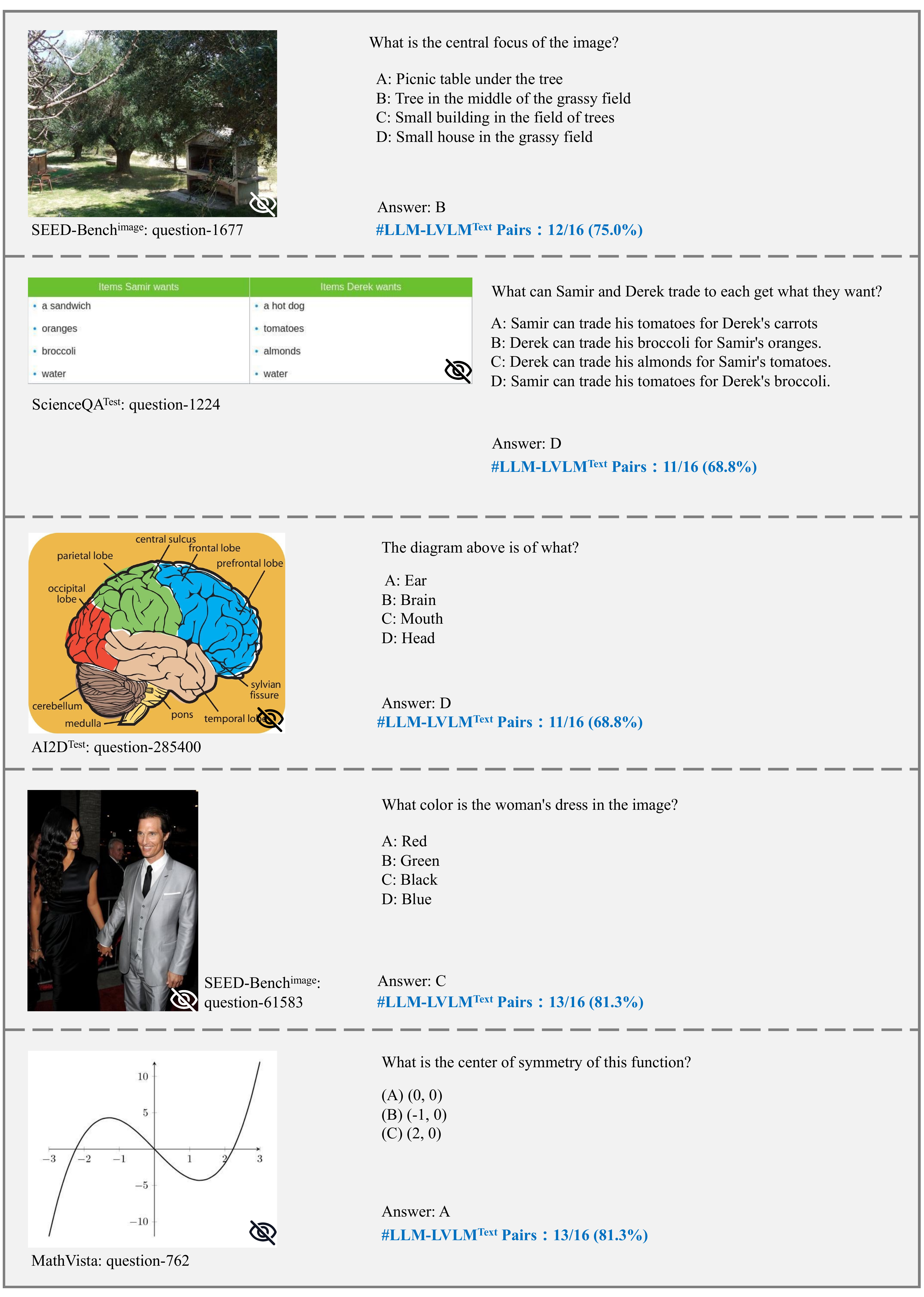}
    \captionsetup{font={footnotesize}}
    \caption{We highlight cases in existing benchmarks where evaluation samples are leaked into LVLMs' multi-modal training data.}
    \label{fig:pairs_case}
\end{figure*}
\clearpage

\section{Detailed Evaluation Results of LVLMs on Six Multi-modal Benchmarks}
\renewcommand{\arraystretch}{1.1}
\begin{table*}[!h]
    \centering
    \footnotesize
    \captionsetup{font={footnotesize}}
    \caption {\textbf{Evaluation of various LVLMs on six popular multi-modal benchmarks.} For the "strategy" column, "LLM" refers to evaluating using the corresponding LLM base of the LVLM, while "LVLM-text" denotes evaluating LVLMs without accessing images. We employ the 0-shot inference strategy for LLMs to align the evaluation protocols of LVLMs. The highest results of the LVLM-text setting across the models are highlighted in {\ul \textbf{bold and underlined.}}}
    \label{tab:detalied_lvlm}
    \scalebox{0.84}{

\begin{tabular}{lllccccccc}
\toprule
\multicolumn{1}{l|}{Model}                                                                                                               & \multicolumn{1}{l|}{Param.}                                        & \multicolumn{1}{l|}{Strategy}                          & MMMU                & MMB                 & ScienceQA           & AI2D                & SEED                & \multicolumn{1}{c|}{MathVista}                                   & Avg.                \\ \midrule
\multicolumn{10}{c}{\textit{Baseline}}                                                                                                                                                                                                                                                                                                                                                                                                                                        \\ \midrule
\multicolumn{1}{l|}{Random Choice}                                                                                                       & \multicolumn{1}{l|}{-}                                             & \multicolumn{1}{l|}{-}                                 & 22.1                & 0.0                & 24.2                & 23.8                & 24.3                & \multicolumn{1}{c|}{17.9}                                        & 18.7                \\ \midrule
\multicolumn{10}{c}{\textit{Closed-source LVLMs and corresponding LLM bases}}                                                                                                                                                                                                                                                                                                                                                                                                 \\ \midrule
\rowcolor[HTML]{E4E4E4} 
\multicolumn{1}{l|}{\cellcolor[HTML]{E4E4E4}}                                                                                            & \multicolumn{1}{l|}{\cellcolor[HTML]{E4E4E4}}                      & \multicolumn{1}{l|}{\cellcolor[HTML]{E4E4E4}LLM}       & 41.2                & 12.2                & 64.3                & 59.7                & 10.1                & \multicolumn{1}{c|}{\cellcolor[HTML]{E4E4E4}24.2}                & 35.3                \\
\rowcolor[HTML]{E4E4E4} 
\multicolumn{1}{l|}{\cellcolor[HTML]{E4E4E4}}                                                                                            & \multicolumn{1}{l|}{\cellcolor[HTML]{E4E4E4}}                      & \multicolumn{1}{l|}{\cellcolor[HTML]{E4E4E4}LVLM-text} & {\ul \textbf{45.1}} & {\ul \textbf{17.6}} & {\ul \textbf{68.2}} & {\ul \textbf{62.5}} & {\ul \textbf{28.4}} & \multicolumn{1}{c|}{\cellcolor[HTML]{E4E4E4}{\ul \textbf{25.4}}} & {\ul \textbf{41.2}} \\
\rowcolor[HTML]{E4E4E4} 
\multicolumn{1}{l|}{\multirow{-3}{*}{\cellcolor[HTML]{E4E4E4}\begin{tabular}[c]{@{}l@{}}GPT4V\cite{gpt4v}\\ (GPT4-Turbo\cite{chatgpt})\end{tabular}}}              & \multicolumn{1}{l|}{\multirow{-3}{*}{\cellcolor[HTML]{E4E4E4}-}}   & \multicolumn{1}{l|}{\cellcolor[HTML]{E4E4E4}LVLM}      & 53.6                & 69.6                & 81.4                & 75.3                & 71.6                & \multicolumn{1}{c|}{\cellcolor[HTML]{E4E4E4}44.7}                & 66.0                \\
\multicolumn{1}{l|}{}                                                                                                                    & \multicolumn{1}{l|}{}                                              & \multicolumn{1}{l|}{LLM}                               & 42.9                & 18.4                & 68.9                & 59.2                & 35.5                & \multicolumn{1}{c|}{23.3}                                        & 41.4                \\
\multicolumn{1}{l|}{}                                                                                                                    & \multicolumn{1}{l|}{}                                              & \multicolumn{1}{l|}{LVLM-text}                         & 39.4                & 16.7                & 66.3                & 54.5                & 27.9                & \multicolumn{1}{c|}{24.5}                                        & 38.2                \\
\multicolumn{1}{l|}{\multirow{-3}{*}{\begin{tabular}[c]{@{}l@{}}GeminiPro-Vision\cite{team2023gemini}\\ (GeminiPro\cite{team2023gemini})\end{tabular}}}                            & \multicolumn{1}{l|}{\multirow{-3}{*}{-}}                           & \multicolumn{1}{l|}{LVLM}                              & 44.4                & 68.1                & 80.6                & 68.0                & 64.3                & \multicolumn{1}{c|}{36.0}                                        & 60.2                \\ \midrule
\multicolumn{10}{c}{\textit{Open-source LVLMs and corresponding LLM bases}}                                                                                                                                                                                                                                                                                                                                                                                                   \\ \midrule
\rowcolor[HTML]{E4E4E4} 
\multicolumn{1}{l|}{\cellcolor[HTML]{E4E4E4}}                                                                                            & \multicolumn{1}{l|}{\cellcolor[HTML]{E4E4E4}}                      & \multicolumn{1}{l|}{\cellcolor[HTML]{E4E4E4}LLM}       & 20.0                & 7.2                 & 47.1                & 38.7                & 26.4                & \multicolumn{1}{c|}{\cellcolor[HTML]{E4E4E4}22.0}                & 26.9                \\
\rowcolor[HTML]{E4E4E4} 
\multicolumn{1}{l|}{\cellcolor[HTML]{E4E4E4}}                                                                                            & \multicolumn{1}{l|}{\cellcolor[HTML]{E4E4E4}}                      & \multicolumn{1}{l|}{\cellcolor[HTML]{E4E4E4}LVLM-text} & 30.0                & 21.0                & 62.3                & 51.9                & 37.2                & \multicolumn{1}{c|}{\cellcolor[HTML]{E4E4E4}23.5}                & 37.7                \\
\rowcolor[HTML]{E4E4E4} 
\multicolumn{1}{l|}{\multirow{-3}{*}{\cellcolor[HTML]{E4E4E4}\begin{tabular}[c]{@{}l@{}}TinyLLaVA\cite{zhou2024tinyllava}\\ (Phi2-2.7B\cite{phi2})\end{tabular}}}           & \multicolumn{1}{l|}{\multirow{-3}{*}{\cellcolor[HTML]{E4E4E4}3B}}  & \multicolumn{1}{l|}{\cellcolor[HTML]{E4E4E4}LVLM}      & 36.0                & 66.9                & 69.1                & 62.4                & 70.1                & \multicolumn{1}{c|}{\cellcolor[HTML]{E4E4E4}28.9}                & 55.6                \\
\multicolumn{1}{l|}{}                                                                                                                    & \multicolumn{1}{l|}{}                                              & \multicolumn{1}{l|}{LLM}                               & 25.7                & 9.5                 & 58.1                & 39.1                & 27.4                & \multicolumn{1}{c|}{21.2}                                        & 30.2                \\
\multicolumn{1}{l|}{}                                                                                                                    & \multicolumn{1}{l|}{}                                              & \multicolumn{1}{l|}{LVLM-text}                         & 33.1                & 23.6                & 67.5                & 55.7                & 38.3                & \multicolumn{1}{c|}{24.2}                                        & 40.4                \\
\multicolumn{1}{l|}{\multirow{-3}{*}{\begin{tabular}[c]{@{}l@{}}Yi-VL\cite{young2024yi}\\ (Yi-6B\cite{young2024yi})\end{tabular}}}                                           & \multicolumn{1}{l|}{\multirow{-3}{*}{6B}}                          & \multicolumn{1}{l|}{LVLM}                              & 38.4                & 69.2                & 72.6                & 59.6                & 67.5                & \multicolumn{1}{c|}{28.0}                                        & 55.9                \\
\rowcolor[HTML]{E4E4E4} 
\multicolumn{1}{l|}{\cellcolor[HTML]{E4E4E4}}                                                                                            & \multicolumn{1}{l|}{\cellcolor[HTML]{E4E4E4}}                      & \multicolumn{1}{l|}{\cellcolor[HTML]{E4E4E4}LLM}       & 29.9                & 10.3                & 58.9                & 42.5                & 32.6                & \multicolumn{1}{c|}{\cellcolor[HTML]{E4E4E4}22.0}                & 32.7                \\
\rowcolor[HTML]{E4E4E4} 
\multicolumn{1}{l|}{\cellcolor[HTML]{E4E4E4}}                                                                                            & \multicolumn{1}{l|}{\cellcolor[HTML]{E4E4E4}}                      & \multicolumn{1}{l|}{\cellcolor[HTML]{E4E4E4}LVLM-text} & 29.9                & 19.5                & 64.1                & 48.7                & 37.5                & \multicolumn{1}{c|}{\cellcolor[HTML]{E4E4E4}20.3}                & 36.7                \\
\rowcolor[HTML]{E4E4E4} 
\multicolumn{1}{l|}{\multirow{-3}{*}{\cellcolor[HTML]{E4E4E4}\begin{tabular}[c]{@{}l@{}}LLaVA-1.5\cite{liu2023improved}\\ (Vicuna-v1.5-7B\cite{chiang2023vicuna})\end{tabular}}}      & \multicolumn{1}{l|}{\multirow{-3}{*}{\cellcolor[HTML]{E4E4E4}7B}}  & \multicolumn{1}{l|}{\cellcolor[HTML]{E4E4E4}LVLM}      & 34.4                & 65.0                & 68.7                & 55.6                & 65.6                & \multicolumn{1}{c|}{\cellcolor[HTML]{E4E4E4}23.6}                & 52.2                \\
\multicolumn{1}{l|}{}                                                                                                                    & \multicolumn{1}{l|}{}                                              & \multicolumn{1}{l|}{LLM}                               & 29.9                & 10.3                & 58.9                & 42.5                & 32.6                & \multicolumn{1}{c|}{22.0}                                        & 32.7                \\
\multicolumn{1}{l|}{}                                                                                                                    & \multicolumn{1}{l|}{}                                              & \multicolumn{1}{l|}{LVLM-text}                         & 31.7                & 20.4                & 65.2                & 49.4                & 37.7                & \multicolumn{1}{c|}{22.7}                                        & 37.9                \\
\multicolumn{1}{l|}{\multirow{-3}{*}{\begin{tabular}[c]{@{}l@{}}ShareGPT4V\cite{chen2023sharegpt4v}\\ (Vicuna-v1.5-7B\cite{chiang2023vicuna})\end{tabular}}}                             & \multicolumn{1}{l|}{\multirow{-3}{*}{7B}}                          & \multicolumn{1}{l|}{LVLM}                              & 35.2                & 69.5                & 69.4                & 57.9                & 69.4                & \multicolumn{1}{c|}{25.7}                                        & 54.5                \\
\rowcolor[HTML]{E4E4E4} 
\multicolumn{1}{l|}{\cellcolor[HTML]{E4E4E4}}                                                                                            & \multicolumn{1}{l|}{\cellcolor[HTML]{E4E4E4}}                      & \multicolumn{1}{l|}{\cellcolor[HTML]{E4E4E4}LLM}       & 32.8                & 8.9                 & 64.0                & 48.3                & 31.9                & \multicolumn{1}{c|}{\cellcolor[HTML]{E4E4E4}18.9}                & 34.1                \\
\rowcolor[HTML]{E4E4E4} 
\multicolumn{1}{l|}{\cellcolor[HTML]{E4E4E4}}                                                                                            & \multicolumn{1}{l|}{\cellcolor[HTML]{E4E4E4}}                      & \multicolumn{1}{l|}{\cellcolor[HTML]{E4E4E4}LVLM-text} & 34.2                & {\ul \textbf{26.2}} & {\ul \textbf{71.9}} & 63.3                & 38.1                & \multicolumn{1}{c|}{\cellcolor[HTML]{E4E4E4}{\ul \textbf{29.4}}} & 43.9                \\
\rowcolor[HTML]{E4E4E4} 
\multicolumn{1}{l|}{\multirow{-3}{*}{\cellcolor[HTML]{E4E4E4}\begin{tabular}[c]{@{}l@{}}InternLM2-XC2\cite{dong2024internlm}\\ (InternLM2-7B\cite{team2023internlm})\end{tabular}}}    & \multicolumn{1}{l|}{\multirow{-3}{*}{\cellcolor[HTML]{E4E4E4}7B}}  & \multicolumn{1}{l|}{\cellcolor[HTML]{E4E4E4}LVLM}      & 41.7                & 79.6                & 96.7                & 81.4                & 74.9                & \multicolumn{1}{c|}{\cellcolor[HTML]{E4E4E4}57.4}                & 72.0                \\
\multicolumn{1}{l|}{}                                                                                                                    & \multicolumn{1}{l|}{}                                              & \multicolumn{1}{l|}{LLM}                               & 19.8                & 8.4                 & 52.7                & 42.6                & 7.6                 & \multicolumn{1}{c|}{20.5}                                        & 25.3                \\
\multicolumn{1}{l|}{}                                                                                                                    & \multicolumn{1}{l|}{}                                              & \multicolumn{1}{l|}{LVLM-text}                         & 24.0                & 8.7                 & 56.7                & 49.0                & 19.5                & \multicolumn{1}{c|}{20.8}                                        & 29.8                \\
\multicolumn{1}{l|}{\multirow{-3}{*}{\begin{tabular}[c]{@{}l@{}}Qwen-VL-Chat\cite{bai2023qwenvl}\\ (Qwen-7B\cite{bai2023qwen})\end{tabular}}}                                  & \multicolumn{1}{l|}{\multirow{-3}{*}{8B}}                          & \multicolumn{1}{l|}{LVLM}                              & 34.0                & 58.3                & 67.7                & 61.3                & 64.0                & \multicolumn{1}{c|}{32.2}                                        & 52.9                \\
\rowcolor[HTML]{E4E4E4} 
\multicolumn{1}{l|}{\cellcolor[HTML]{E4E4E4}}                                                                                            & \multicolumn{1}{l|}{\cellcolor[HTML]{E4E4E4}}                      & \multicolumn{1}{l|}{\cellcolor[HTML]{E4E4E4}LLM}       & 21.6                & 8.4                 & 56.3                & 38.1                & 13.4                & \multicolumn{1}{c|}{\cellcolor[HTML]{E4E4E4}20.6}                & 26.4                \\
\rowcolor[HTML]{E4E4E4} 
\multicolumn{1}{l|}{\cellcolor[HTML]{E4E4E4}}                                                                                            & \multicolumn{1}{l|}{\cellcolor[HTML]{E4E4E4}}                      & \multicolumn{1}{l|}{\cellcolor[HTML]{E4E4E4}LVLM-text} & 32.2                & 23.9                & 67.1                & 53.0                & 36.5                & \multicolumn{1}{c|}{\cellcolor[HTML]{E4E4E4}23.9}                & 39.4                \\
\rowcolor[HTML]{E4E4E4} 
\multicolumn{1}{l|}{\multirow{-3}{*}{\cellcolor[HTML]{E4E4E4}\begin{tabular}[c]{@{}l@{}}Deepseek-VL\cite{lu2024deepseek}\\ (Deepseek-7B\cite{bi2024deepseek})\end{tabular}}}       & \multicolumn{1}{l|}{\multirow{-3}{*}{\cellcolor[HTML]{E4E4E4}8B}}  & \multicolumn{1}{l|}{\cellcolor[HTML]{E4E4E4}LVLM}      & 35.4                & 73.5                & 81.4                & 64.6                & 70.2                & \multicolumn{1}{c|}{\cellcolor[HTML]{E4E4E4}35.3}                & 60.1                \\
\multicolumn{1}{l|}{}                                                                                                                    & \multicolumn{1}{l|}{}                                              & \multicolumn{1}{l|}{LLM}                               & 19.8                & 8.4                 & 52.7                & 42.6                & 7.6                 & \multicolumn{1}{c|}{20.5}                                        & 25.3                \\
\multicolumn{1}{l|}{}                                                                                                                    & \multicolumn{1}{l|}{}                                              & \multicolumn{1}{l|}{LVLM-text}                         & 32.4                & 15.6                & 71.1                & 56.8                & 36.1                & \multicolumn{1}{c|}{25.0}                                        & 39.5                \\
\multicolumn{1}{l|}{\multirow{-3}{*}{\begin{tabular}[c]{@{}l@{}}Monkey-Chat\cite{li2023monkey}\\ (Qwen-7B\cite{bai2023qwen})\end{tabular}}}                                   & \multicolumn{1}{l|}{\multirow{-3}{*}{10B}}                         & \multicolumn{1}{l|}{LVLM}                              & 37.1                & 71.0                & 82.4                & 68.5                & 69.1                & \multicolumn{1}{c|}{34.0}                                        & 60.4                \\
\rowcolor[HTML]{E4E4E4} 
\multicolumn{1}{l|}{\cellcolor[HTML]{E4E4E4}}                                                                                            & \multicolumn{1}{l|}{\cellcolor[HTML]{E4E4E4}}                      & \multicolumn{1}{l|}{\cellcolor[HTML]{E4E4E4}LLM}       & 28.3                & 11.6                & 59.5                & 45.0                & 26.3                & \multicolumn{1}{c|}{\cellcolor[HTML]{E4E4E4}19.6}                & 31.7                \\
\rowcolor[HTML]{E4E4E4} 
\multicolumn{1}{l|}{\cellcolor[HTML]{E4E4E4}}                                                                                            & \multicolumn{1}{l|}{\cellcolor[HTML]{E4E4E4}}                      & \multicolumn{1}{l|}{\cellcolor[HTML]{E4E4E4}LVLM-text} & 26.0                & 21.4                & 66.5                & 52.2                & 37.0                & \multicolumn{1}{c|}{\cellcolor[HTML]{E4E4E4}21.1}                & 37.4                \\
\rowcolor[HTML]{E4E4E4} 
\multicolumn{1}{l|}{\multirow{-3}{*}{\cellcolor[HTML]{E4E4E4}\begin{tabular}[c]{@{}l@{}}LLaVA-1.5\cite{liu2023improved}\\ (Vicuna-v1.5-13B\cite{chiang2023vicuna})\end{tabular}}}     & \multicolumn{1}{l|}{\multirow{-3}{*}{\cellcolor[HTML]{E4E4E4}13B}} & \multicolumn{1}{l|}{\cellcolor[HTML]{E4E4E4}LVLM}      & 35.6                & 68.6                & 72.2                & 60.8                & 68.1                & \multicolumn{1}{c|}{\cellcolor[HTML]{E4E4E4}26.4}                & 55.3                \\
\multicolumn{1}{l|}{}                                                                                                                    & \multicolumn{1}{l|}{}                                              & \multicolumn{1}{l|}{LLM}                               & 29.9                & 10.3                & 58.9                & 42.5                & 32.6                & \multicolumn{1}{c|}{22.0}                                        & 32.7                \\
\multicolumn{1}{l|}{}                                                                                                                    & \multicolumn{1}{l|}{}                                              & \multicolumn{1}{l|}{LVLM-text}                         & 30.1                & 15.5                & 54.6                & 52.5                & 36.7                & \multicolumn{1}{c|}{25.0}                                        & 35.7                \\
\multicolumn{1}{l|}{\multirow{-3}{*}{\begin{tabular}[c]{@{}l@{}}CogVLM-Chat\cite{wang2023cogvlm}\\ (Vicuna-v1.5-7B\cite{chiang2023vicuna})\end{tabular}}}                            & \multicolumn{1}{l|}{\multirow{-3}{*}{17B}}                         & \multicolumn{1}{l|}{LVLM}                              & 34.2                & 63.4                & 66.3                & 63.3                & 68.7                & \multicolumn{1}{c|}{34.7}                                        & 55.1                \\
\rowcolor[HTML]{E4E4E4} 
\multicolumn{1}{l|}{\cellcolor[HTML]{E4E4E4}}                                                                                            & \multicolumn{1}{l|}{\cellcolor[HTML]{E4E4E4}}                      & \multicolumn{1}{l|}{\cellcolor[HTML]{E4E4E4}LLM}       & 37.1                & 10.5                & 53.6                & 57.3                & 37.3                & \multicolumn{1}{c|}{\cellcolor[HTML]{E4E4E4}21.7}                & 36.3                \\
\rowcolor[HTML]{E4E4E4} 
\multicolumn{1}{l|}{\cellcolor[HTML]{E4E4E4}}                                                                                            & \multicolumn{1}{l|}{\cellcolor[HTML]{E4E4E4}}                      & \multicolumn{1}{l|}{\cellcolor[HTML]{E4E4E4}LVLM-text} & 37.3                & 23.2                & 68.6                & 59.9                & {\ul \textbf{41.0}} & \multicolumn{1}{c|}{\cellcolor[HTML]{E4E4E4}22.7}                & 42.1                \\
\rowcolor[HTML]{E4E4E4} 
\multicolumn{1}{l|}{\multirow{-3}{*}{\cellcolor[HTML]{E4E4E4}\begin{tabular}[c]{@{}l@{}}Yi-VL\cite{young2024yi}\\ (Yi-34B\cite{young2024yi})\end{tabular}}}                  & \multicolumn{1}{l|}{\multirow{-3}{*}{\cellcolor[HTML]{E4E4E4}34B}} & \multicolumn{1}{l|}{\cellcolor[HTML]{E4E4E4}LVLM}      & 43.2                & 71.5                & 75.3                & 65.9                & 68.1                & \multicolumn{1}{c|}{\cellcolor[HTML]{E4E4E4}25.6}                & 58.3                \\
\multicolumn{1}{l|}{}                                                                                                                    & \multicolumn{1}{l|}{}                                              & \multicolumn{1}{l|}{LLM}                               & 37.6                & 20.1                & 69.4                & 60.2                & 35.0                & \multicolumn{1}{c|}{17.9}                                        & 37.2                \\
\multicolumn{1}{l|}{}                                                                                                                    & \multicolumn{1}{l|}{}                                              & \multicolumn{1}{l|}{LVLM-text}                         & 40.4                & 24.9                & 70.9                & 65.8                & 41.7                & \multicolumn{1}{c|}{22.2}                                        & 44.3                \\
\multicolumn{1}{l|}{\multirow{-3}{*}{\begin{tabular}[c]{@{}l@{}}LLaVA-Next\cite{liu2024llavanext}\\ (NH2-Yi-34B\cite{nousyi34b})\end{tabular}}}                                 & \multicolumn{1}{l|}{\multirow{-3}{*}{34B}}                         & \multicolumn{1}{l|}{LVLM}                              & 47.0                & 79.6                & 82.1                & 78.6                & 75.8                & \multicolumn{1}{c|}{38.7}                                        & 67.0                \\
\rowcolor[HTML]{E4E4E4} 
\multicolumn{1}{l|}{\cellcolor[HTML]{E4E4E4}}                                                                                            & \multicolumn{1}{l|}{\cellcolor[HTML]{E4E4E4}}                      & \multicolumn{1}{l|}{\cellcolor[HTML]{E4E4E4}LLM}       & 37.6                & 20.1                & 69.4                & 60.2                & 35.0                & \multicolumn{1}{c|}{\cellcolor[HTML]{E4E4E4}17.9}                & 40.0                \\
\rowcolor[HTML]{E4E4E4} 
\multicolumn{1}{l|}{\cellcolor[HTML]{E4E4E4}}                                                                                            & \multicolumn{1}{l|}{\cellcolor[HTML]{E4E4E4}}                      & \multicolumn{1}{l|}{\cellcolor[HTML]{E4E4E4}LVLM-text} & 41.7                & 23.9                & 70.3                & {\ul \textbf{65.0}} & 40.5                & \multicolumn{1}{c|}{\cellcolor[HTML]{E4E4E4}24.0}                & {\ul \textbf{44.2}} \\
\rowcolor[HTML]{E4E4E4} 
\multicolumn{1}{l|}{\multirow{-3}{*}{\cellcolor[HTML]{E4E4E4}\begin{tabular}[c]{@{}l@{}}InternVL-Chat-v1.2\cite{chen2023internvl}\\ (NH2-Yi-34B\cite{nousyi34b})\end{tabular}}} & \multicolumn{1}{l|}{\multirow{-3}{*}{\cellcolor[HTML]{E4E4E4}40B}} & \multicolumn{1}{l|}{\cellcolor[HTML]{E4E4E4}LVLM}      & 49.1                & 82.4                & 82.5                & 78.5                & 75.4                & \multicolumn{1}{c|}{\cellcolor[HTML]{E4E4E4}47.7}                & 69.3                \\
\multicolumn{1}{l|}{}                                                                                                                    & \multicolumn{1}{l|}{}                                              & \multicolumn{1}{l|}{LLM}                               & 25.7                & 8.6                 & 57.2                & 48.7                & 13.5                & \multicolumn{1}{c|}{23.4}                                        & 29.5                \\
\multicolumn{1}{l|}{}                                                                                                                    & \multicolumn{1}{l|}{}                                              & \multicolumn{1}{l|}{LVLM-text}                         & {\ul \textbf{43.6}} & 20.5                & 68.4                & 61.1                & 39.9                & \multicolumn{1}{c|}{28.4}                                        & 43.7                \\
\multicolumn{1}{l|}{\multirow{-3}{*}{\begin{tabular}[c]{@{}l@{}}Sphinx-X-MoE\cite{gao2024sphinx}\\ (Mixtral-8x7B\cite{jiang2024mixtral})\end{tabular}}}                             & \multicolumn{1}{l|}{\multirow{-3}{*}{57B}}                         & \multicolumn{1}{l|}{LVLM}                              & 44.8                & 69.2                & 72.2                & 65.0                & 71.1                & \multicolumn{1}{c|}{38.1}                                        & 60.1                \\ \bottomrule
\end{tabular}
}
\end{table*}

\clearpage

{
\small
\bibliographystyle{abbrv}
\bibliography{main}

\begin{thebibliography}{10}

\bibitem{bai2023qwen}
J.~Bai, S.~Bai, Y.~Chu, Z.~Cui, K.~Dang, X.~Deng, Y.~Fan, W.~Ge, Y.~Han, F.~Huang, et~al.
\newblock Qwen technical report.
\newblock {\em arXiv preprint arXiv:2309.16609}, 2023.

\bibitem{bai2023qwenvl}
J.~Bai, S.~Bai, S.~Yang, S.~Wang, S.~Tan, P.~Wang, J.~Lin, C.~Zhou, and J.~Zhou.
\newblock Qwen-vl: A frontier large vision-language model with versatile abilities.
\newblock {\em arXiv preprint arXiv:2308.12966}, 2023.

\bibitem{bi2024deepseek}
X.~Bi, D.~Chen, G.~Chen, S.~Chen, D.~Dai, C.~Deng, H.~Ding, K.~Dong, Q.~Du, Z.~Fu, et~al.
\newblock Deepseek llm: Scaling open-source language models with longtermism.
\newblock {\em arXiv preprint arXiv:2401.02954}, 2024.

\bibitem{brown2020language}
T.~Brown, B.~Mann, N.~Ryder, M.~Subbiah, J.~D. Kaplan, P.~Dhariwal, A.~Neelakantan, P.~Shyam, G.~Sastry, A.~Askell, et~al.
\newblock Language models are few-shot learners.
\newblock {\em Advances in neural information processing systems}, 33:1877--1901, 2020.

\bibitem{chen2023sharegpt4v}
L.~Chen, J.~Li, X.~Dong, P.~Zhang, C.~He, J.~Wang, F.~Zhao, and D.~Lin.
\newblock Sharegpt4v: Improving large multi-modal models with better captions.
\newblock {\em arXiv preprint arXiv:2311.12793}, 2023.

\bibitem{chen2023internvl}
Z.~Chen, J.~Wu, W.~Wang, W.~Su, G.~Chen, S.~Xing, Z.~Muyan, Q.~Zhang, X.~Zhu, L.~Lu, et~al.
\newblock Internvl: Scaling up vision foundation models and aligning for generic visual-linguistic tasks.
\newblock {\em arXiv preprint arXiv:2312.14238}, 2023.

\bibitem{cheng2023can}
S.~Cheng, Z.~Guo, J.~Wu, K.~Fang, P.~Li, H.~Liu, and Y.~Liu.
\newblock Can vision-language models think from a first-person perspective?
\newblock {\em arXiv preprint arXiv:2311.15596}, 2023.

\bibitem{chiang2023vicuna}
W.-L. Chiang, Z.~Li, Z.~Lin, Y.~Sheng, Z.~Wu, H.~Zhang, L.~Zheng, S.~Zhuang, Y.~Zhuang, J.~E. Gonzalez, et~al.
\newblock Vicuna: An open-source chatbot impressing gpt-4 with 90\%* chatgpt quality.
\newblock {\em See https://vicuna. lmsys. org (accessed 14 April 2023)}, 2023.

\bibitem{chowdhery2022palm}
A.~Chowdhery, S.~Narang, J.~Devlin, M.~Bosma, G.~Mishra, A.~Roberts, P.~Barham, H.~W. Chung, C.~Sutton, S.~Gehrmann, et~al.
\newblock Palm: Scaling language modeling with pathways.
\newblock {\em arXiv preprint arXiv:2204.02311}, 2022.

\bibitem{2023opencompass}
O.~Contributors.
\newblock Opencompass: A universal evaluation platform for foundation models.
\newblock \url{https://github.com/open-compass/opencompass}, 2023.

\bibitem{instructblip}
W.~Dai, J.~Li, D.~Li, A.~M.~H. Tiong, J.~Zhao, W.~Wang, B.~Li, P.~Fung, and S.~Hoi.
\newblock Instructblip: Towards general-purpose vision-language models with instruction tuning, 2023.

\bibitem{dong2024internlm}
X.~Dong, P.~Zhang, Y.~Zang, Y.~Cao, B.~Wang, L.~Ouyang, X.~Wei, S.~Zhang, H.~Duan, M.~Cao, et~al.
\newblock Internlm-xcomposer2: Mastering free-form text-image composition and comprehension in vision-language large model.
\newblock {\em arXiv preprint arXiv:2401.16420}, 2024.

\bibitem{du2021glm}
Z.~Du, Y.~Qian, X.~Liu, M.~Ding, J.~Qiu, Z.~Yang, and J.~Tang.
\newblock Glm: General language model pretraining with autoregressive blank infilling.
\newblock {\em arXiv preprint arXiv:2103.10360}, 2021.

\bibitem{fu2023mme}
C.~Fu, P.~Chen, Y.~Shen, Y.~Qin, M.~Zhang, X.~Lin, Z.~Qiu, W.~Lin, J.~Yang, X.~Zheng, K.~Li, X.~Sun, and R.~Ji.
\newblock Mme: A comprehensive evaluation benchmark for multimodal large language models.
\newblock {\em arXiv preprint arXiv:2306.13394}, 2023.

\bibitem{gao2024sphinx}
P.~Gao, R.~Zhang, C.~Liu, L.~Qiu, S.~Huang, W.~Lin, S.~Zhao, S.~Geng, Z.~Lin, P.~Jin, et~al.
\newblock Sphinx-x: Scaling data and parameters for a family of multi-modal large language models.
\newblock {\em arXiv preprint arXiv:2402.05935}, 2024.

\bibitem{goyal2017making}
Y.~Goyal, T.~Khot, D.~Summers-Stay, D.~Batra, and D.~Parikh.
\newblock Making the v in vqa matter: Elevating the role of image understanding in visual question answering.
\newblock In {\em Proceedings of the IEEE conference on computer vision and pattern recognition}, pages 6904--6913, 2017.

\bibitem{jia2021scaling}
C.~Jia, Y.~Yang, Y.~Xia, Y.-T. Chen, Z.~Parekh, H.~Pham, Q.~Le, Y.-H. Sung, Z.~Li, and T.~Duerig.
\newblock Scaling up visual and vision-language representation learning with noisy text supervision.
\newblock In {\em International conference on machine learning}, pages 4904--4916. PMLR, 2021.

\bibitem{jiang2023mistral}
A.~Q. Jiang, A.~Sablayrolles, A.~Mensch, C.~Bamford, D.~S. Chaplot, D.~d.~l. Casas, F.~Bressand, G.~Lengyel, G.~Lample, L.~Saulnier, et~al.
\newblock Mistral 7b.
\newblock {\em arXiv preprint arXiv:2310.06825}, 2023.

\bibitem{jiang2024mixtral}
A.~Q. Jiang, A.~Sablayrolles, A.~Roux, A.~Mensch, B.~Savary, C.~Bamford, D.~S. Chaplot, D.~d.~l. Casas, E.~B. Hanna, F.~Bressand, et~al.
\newblock Mixtral of experts.
\newblock {\em arXiv preprint arXiv:2401.04088}, 2024.

\bibitem{Kembhavi2016ADI}
A.~Kembhavi, M.~Salvato, E.~Kolve, M.~Seo, H.~Hajishirzi, and A.~Farhadi.
\newblock A diagram is worth a dozen images.
\newblock {\em ArXiv}, abs/1603.07396, 2016.

\bibitem{li2023seed}
B.~Li, R.~Wang, G.~Wang, Y.~Ge, Y.~Ge, and Y.~Shan.
\newblock Seed-bench: Benchmarking multimodal llms with generative comprehension.
\newblock {\em arXiv preprint arXiv:2307.16125}, 2023.

\bibitem{li2023blip}
J.~Li, D.~Li, S.~Savarese, and S.~Hoi.
\newblock Blip-2: Bootstrapping language-image pre-training with frozen image encoders and large language models.
\newblock {\em arXiv preprint arXiv:2301.12597}, 2023.

\bibitem{li2023monkey}
Z.~Li, B.~Yang, Q.~Liu, Z.~Ma, S.~Zhang, J.~Yang, Y.~Sun, Y.~Liu, and X.~Bai.
\newblock Monkey: Image resolution and text label are important things for large multi-modal models.
\newblock {\em arXiv preprint arXiv:2311.06607}, 2023.

\bibitem{liu2023improved}
H.~Liu, C.~Li, Y.~Li, and Y.~J. Lee.
\newblock Improved baselines with visual instruction tuning.
\newblock {\em arXiv preprint arXiv:2310.03744}, 2023.

\bibitem{liu2024llavanext}
H.~Liu, C.~Li, Y.~Li, B.~Li, Y.~Zhang, S.~Shen, and Y.~J. Lee.
\newblock Llava-next: Improved reasoning, ocr, and world knowledge, January 2024.

\bibitem{liu2023visual}
H.~Liu, C.~Li, Q.~Wu, and Y.~J. Lee.
\newblock Visual instruction tuning.
\newblock {\em arXiv preprint arXiv:2304.08485}, 2023.

\bibitem{liu2023mmbench}
Y.~Liu, H.~Duan, Y.~Zhang, B.~Li, S.~Zhang, W.~Zhao, Y.~Yuan, J.~Wang, C.~He, Z.~Liu, et~al.
\newblock Mmbench: Is your multi-modal model an all-around player?
\newblock {\em arXiv preprint arXiv:2307.06281}, 2023.

\bibitem{lu2024deepseek}
H.~Lu, W.~Liu, B.~Zhang, B.~Wang, K.~Dong, B.~Liu, J.~Sun, T.~Ren, Z.~Li, Y.~Sun, et~al.
\newblock Deepseek-vl: Towards real-world vision-language understanding.
\newblock {\em arXiv preprint arXiv:2403.05525}, 2024.

\bibitem{lu2023mathvista}
P.~Lu, H.~Bansal, T.~Xia, J.~Liu, C.~Li, H.~Hajishirzi, H.~Cheng, K.-W. Chang, M.~Galley, and J.~Gao.
\newblock Mathvista: Evaluating mathematical reasoning of foundation models in visual contexts.
\newblock {\em arXiv preprint arXiv:2310.02255}, 2023.

\bibitem{lu2022learn}
P.~Lu, S.~Mishra, T.~Xia, L.~Qiu, K.-W. Chang, S.-C. Zhu, O.~Tafjord, P.~Clark, and A.~Kalyan.
\newblock Learn to explain: Multimodal reasoning via thought chains for science question answering.
\newblock {\em Advances in Neural Information Processing Systems}, 35:2507--2521, 2022.

\bibitem{luo2023cheap}
G.~Luo, Y.~Zhou, T.~Ren, S.~Chen, X.~Sun, and R.~Ji.
\newblock Cheap and quick: Efficient vision-language instruction tuning for large language models.
\newblock {\em arXiv preprint arXiv:2305.15023}, 2023.

\bibitem{phi2}
Microsoft.
\newblock Phi2: The surprising power of small language models.
\newblock \url{https://www.microsoft.com/en-us/research/blog/phi-2-the-surprising-power-of-small-language-models/}, 2023.

\bibitem{nousyi34b}
NousResearch.
\newblock Nous-hermes-2-yi-34b.
\newblock \url{https://huggingface.co/NousResearch/Nous-Hermes-2-Yi-34B}, 2023.

\bibitem{chatgpt}
OpenAI.
\newblock Chatgpt.
\newblock \url{https://chat.openai.com/}, 2023.

\bibitem{gpt4v}
OpenAI.
\newblock Gpt-4v(ision) system card.
\newblock \url{https://cdn.openai.com/papers/GPTV_System_Card.pdf}, 2023.

\bibitem{ouyang2022training}
L.~Ouyang, J.~Wu, X.~Jiang, D.~Almeida, C.~Wainwright, P.~Mishkin, C.~Zhang, S.~Agarwal, K.~Slama, A.~Ray, et~al.
\newblock Training language models to follow instructions with human feedback.
\newblock {\em Advances in Neural Information Processing Systems}, 35:27730--27744, 2022.

\bibitem{radford2021learning}
A.~Radford, J.~W. Kim, C.~Hallacy, A.~Ramesh, G.~Goh, S.~Agarwal, G.~Sastry, A.~Askell, P.~Mishkin, J.~Clark, et~al.
\newblock Learning transferable visual models from natural language supervision.
\newblock In {\em International conference on machine learning}, pages 8748--8763. PMLR, 2021.

\bibitem{schwenk2022okvqa}
D.~Schwenk, A.~Khandelwal, C.~Clark, K.~Marino, and R.~Mottaghi.
\newblock A-okvqa: A benchmark for visual question answering using world knowledge.
\newblock In {\em European Conference on Computer Vision}, pages 146--162. Springer, 2022.

\bibitem{sharma2018conceptual}
P.~Sharma, N.~Ding, S.~Goodman, and R.~Soricut.
\newblock Conceptual captions: A cleaned, hypernymed, image alt-text dataset for automatic image captioning.
\newblock In {\em Proceedings of the 56th Annual Meeting of the Association for Computational Linguistics (Volume 1: Long Papers)}, pages 2556--2565, 2018.

\bibitem{taud2018multilayer}
H.~Taud and J.-F. Mas.
\newblock Multilayer perceptron (mlp).
\newblock {\em Geomatic approaches for modeling land change scenarios}, pages 451--455, 2018.

\bibitem{team2023gemini}
G.~Team, R.~Anil, S.~Borgeaud, Y.~Wu, J.-B. Alayrac, J.~Yu, R.~Soricut, J.~Schalkwyk, A.~M. Dai, A.~Hauth, et~al.
\newblock Gemini: a family of highly capable multimodal models.
\newblock {\em arXiv preprint arXiv:2312.11805}, 2023.

\bibitem{team2023internlm}
I.~Team.
\newblock Internlm: A multilingual language model with progressively enhanced capabilities, 2023.

\bibitem{touvron2023llama}
H.~Touvron, L.~Martin, K.~Stone, P.~Albert, A.~Almahairi, Y.~Babaei, N.~Bashlykov, S.~Batra, P.~Bhargava, S.~Bhosale, et~al.
\newblock Llama 2: Open foundation and fine-tuned chat models.
\newblock {\em arXiv preprint arXiv:2307.09288}, 2023.

\bibitem{wang2023see}
J.~Wang, L.~Meng, Z.~Weng, B.~He, Z.~Wu, and Y.-G. Jiang.
\newblock To see is to believe: Prompting gpt-4v for better visual instruction tuning.
\newblock {\em arXiv preprint arXiv:2311.07574}, 2023.

\bibitem{wang2023cogvlm}
W.~Wang, Q.~Lv, W.~Yu, W.~Hong, J.~Qi, Y.~Wang, J.~Ji, Z.~Yang, L.~Zhao, X.~Song, et~al.
\newblock Cogvlm: Visual expert for pretrained language models.
\newblock {\em arXiv preprint arXiv:2311.03079}, 2023.

\bibitem{wu2023q}
H.~Wu, Z.~Zhang, E.~Zhang, C.~Chen, L.~Liao, A.~Wang, C.~Li, W.~Sun, Q.~Yan, G.~Zhai, et~al.
\newblock Q-bench: A benchmark for general-purpose foundation models on low-level vision.
\newblock {\em arXiv preprint arXiv:2309.14181}, 2023.

\bibitem{yang2023baichuan}
A.~Yang, B.~Xiao, B.~Wang, B.~Zhang, C.~Yin, C.~Lv, D.~Pan, D.~Wang, D.~Yan, F.~Yang, et~al.
\newblock Baichuan 2: Open large-scale language models.
\newblock {\em arXiv preprint arXiv:2309.10305}, 2023.

\bibitem{ye2023mplug}
Q.~Ye, H.~Xu, G.~Xu, J.~Ye, M.~Yan, Y.~Zhou, J.~Wang, A.~Hu, P.~Shi, Y.~Shi, et~al.
\newblock mplug-owl: Modularization empowers large language models with multimodality.
\newblock {\em arXiv preprint arXiv:2304.14178}, 2023.

\bibitem{young2024yi}
A.~Young, B.~Chen, C.~Li, C.~Huang, G.~Zhang, G.~Zhang, H.~Li, J.~Zhu, J.~Chen, J.~Chang, et~al.
\newblock Yi: Open foundation models by 01. ai.
\newblock {\em arXiv preprint arXiv:2403.04652}, 2024.

\bibitem{yu2023mm}
W.~Yu, Z.~Yang, L.~Li, J.~Wang, K.~Lin, Z.~Liu, X.~Wang, and L.~Wang.
\newblock Mm-vet: Evaluating large multimodal models for integrated capabilities.
\newblock {\em arXiv preprint arXiv:2308.02490}, 2023.

\bibitem{yue2023mmmu}
X.~Yue, Y.~Ni, K.~Zhang, T.~Zheng, R.~Liu, G.~Zhang, S.~Stevens, D.~Jiang, W.~Ren, Y.~Sun, et~al.
\newblock Mmmu: A massive multi-discipline multimodal understanding and reasoning benchmark for expert agi.
\newblock {\em arXiv preprint arXiv:2311.16502}, 2023.

\bibitem{zhang2023internlm}
P.~Zhang, X.~D.~B. Wang, Y.~Cao, C.~Xu, L.~Ouyang, Z.~Zhao, S.~Ding, S.~Zhang, H.~Duan, H.~Yan, et~al.
\newblock Internlm-xcomposer: A vision-language large model for advanced text-image comprehension and composition.
\newblock {\em arXiv preprint arXiv:2309.15112}, 2023.

\bibitem{zhou2024tinyllava}
B.~Zhou, Y.~Hu, X.~Weng, J.~Jia, J.~Luo, X.~Liu, J.~Wu, and L.~Huang.
\newblock Tinyllava: A framework of small-scale large multimodal models.
\newblock {\em arXiv preprint arXiv:2402.14289}, 2024.

\bibitem{zhu2023minigpt}
D.~Zhu, J.~Chen, X.~Shen, X.~Li, and M.~Elhoseiny.
\newblock Minigpt-4: Enhancing vision-language understanding with advanced large language models.
\newblock {\em arXiv preprint arXiv:2304.10592}, 2023.

\end{thebibliography}
}

\end{document}